%% file: main.tex
\documentclass[journal]{IEEEtran}
\usepackage[utf8]{inputenc} % Switch to utf8 encoding
\usepackage{babel}
\usepackage{float}
\usepackage{algorithm}
\usepackage{algpseudocode}
\usepackage{amssymb}
\usepackage{booktabs}
\usepackage{multirow}
\usepackage{rotating}
\usepackage{graphicx}
\usepackage{subcaption}
\usepackage{setspace}
\usepackage{comment}
\usepackage{threeparttable}
\usepackage{mathrsfs}
\usepackage{array}
\usepackage{pifont}
\usepackage{array}
\usepackage{url}
\usepackage{cite}

\usepackage[hidelinks]{hyperref}

\usepackage{titlesec} % 用于定制appendix标题

\usepackage{comment}
\usepackage{bbm}

\makeatletter

\usepackage{diagbox}

\usepackage{amsmath}

\usepackage{tikz,xcolor}

\makeatother

\begin{document}
\title{Systematic Benchmarking of SUMO Against Data-Driven Traffic Simulators}
\author{
\IEEEauthorblockN{Erdao Liang\\}
\IEEEauthorblockA{
University of Michigan\\
Ann Arbor, USA\\
erdao@umich.edu}
}

\maketitle

% % The code and enhanced traffic signal data are open-sourced here: \url{http://www.tbd}

% \hfill\break%
% \noindent\textit{Keywords}: Waymo Open Motion Dataset, Autonomous Vehicles

\begin{abstract}
This paper presents a systematic benchmarking of the model-based microscopic traffic simulator SUMO against state-of-the-art data-driven traffic simulators using large-scale real-world datasets. Using the Waymo Open Motion Dataset (WOMD) and the Waymo Open Sim Agents Challenge (WOSAC), we evaluate SUMO under both short-horizon (8s) and long-horizon (60s) closed-loop simulation settings. To enable scalable evaluation, we develop Waymo2SUMO, an automated pipeline that converts WOMD scenarios into SUMO simulations. On the WOSAC benchmark, SUMO achieves a realism meta metric of 0.653 while requiring fewer than 100 tunable parameters. Extended rollouts show that SUMO maintains low collision and offroad rates and exhibits stronger long-horizon stability than representative data-driven simulators. These results highlight complementary strengths of model-based and data-driven approaches for autonomous driving simulation and benchmarking. Source code is available at \url{https://github.com/LuminousLamp/SUMO-Benchmark}.

\end{abstract}


\input{sections/intro}
\input{sections/related}

\input{sections/method}

\input{sections/results}

\input{sections/discussions}

\input{sections/conclusions}

\bibliographystyle{IEEEtran}
\bibliography{ref}

\newpage

\end{document}

%% file: sections/intro.tex
\begin{figure*}[t]
    \centering
    \includegraphics[width=0.9\textwidth]{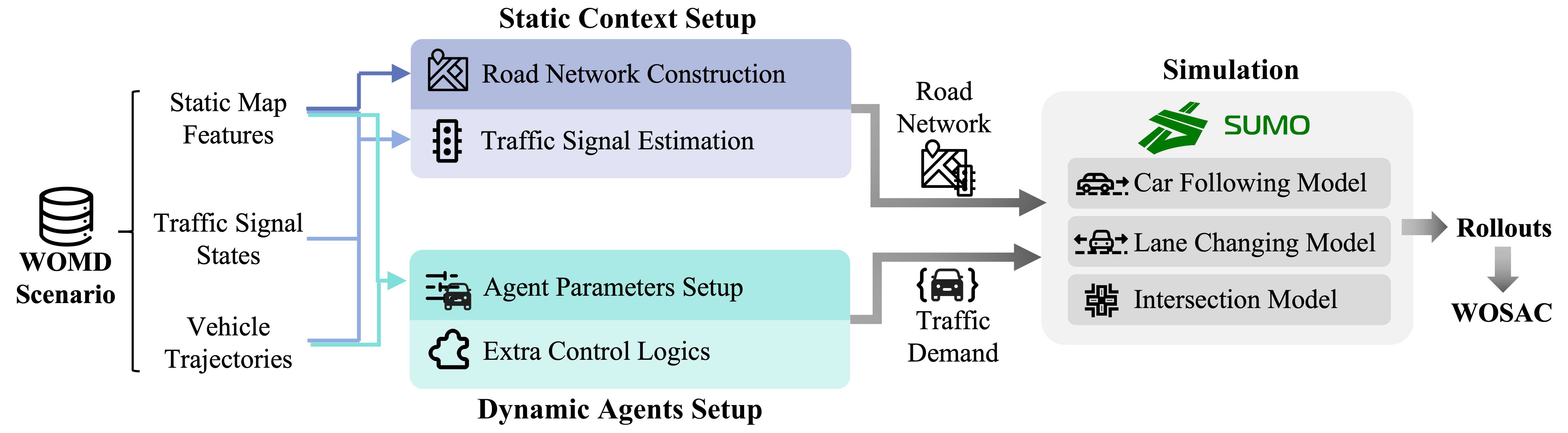}
    \caption{The overall framework.}
    \label{fig:framework}
\end{figure*}

%%%%%%%%%%%%%%%%%%%%%%%%%%%%%%%%%%%%%%%%%%%%%%%%%%%%%%%%%%%%%%%%%%%%%%%%%%
\section{Introduction}

Simulation is a cornerstone for the development and deployment of autonomous vehicle (AV) technologies. It plays a critical role across stakeholders: Developers leverage simulations to train and optimize AV algorithms, third-party testers use them for safety verification and validation, and policymakers rely on simulation insights to establish standards and regulations~\cite{koopman2018framework}. As AV technology advances toward widespread deployment, such as Waymo's operations in San Francisco, Los Angeles, and Phoenix~\cite{waymo_service_areas}, the demand for high-fidelity, efficient, and scalable simulation tools has become more critical than ever.

Broadly, traffic simulation spans model-based and data-driven approaches. Model-based microscopic simulators generate interactive rollouts on structured road networks using parameterized behavior models~\cite{krajzewicz2012sumo,fellendorf2010vissim,casas2010aimsun}, and they remain a practical choice when interpretability and computational efficiency are important. In parallel, end-to-end driving simulators provide richer sensor and environment modeling for testing autonomy stacks under configurable conditions~\cite{dosovitskiy2017carla,rong2020lgsvl,shah2017airsim}. More recently, data-driven simulators have emerged that learn multi-agent behaviors directly from large-scale driving data, showing strong realism for short-horizon motion generation in both industrial and academic settings~\cite{waymo_simulationcity,nvidia_av_simulation,waabi_world,suo2021trafficsim,xu2022bits,igl2022symphony,zhang2023trafficbots,wu2024smart}. Across these paradigms, simulator fidelity is paramount because it directly affects the credibility of training, testing, and scenario-based validation~\cite{zhong2021survey, li2024choose}.

Despite rapid progress, systematic benchmarks that transparently compare model-based simulators (e.g., SUMO) with state-of-the-art data-driven simulators remain limited. Motivated by this gap, we aim to provide an objective evaluation of SUMO on large-scale real-world datasets and to compare it against leading data-driven approaches. Recent public benchmarks and toolkits have begun to standardize closed-loop evaluation for interactive agents, including WOSAC~\cite{montali2024waymo-WOSAC-benchmark} and accelerator-oriented frameworks such as Waymax~\cite{gulino2023waymax}, enabling more reproducible comparisons.

To address these challenges, we leverage the Waymo Open Motion Dataset (WOMD)~\cite{ettinger2021large-womd} and its associated benchmark, the Waymo Open Sim Agent Challenge (WOSAC)~\cite{montali2024waymo-WOSAC-benchmark}. WOMD provides large-scale real-world traffic data with diverse road geometries and agent types, and WOSAC tasks participants with generating 8-second simulations based on 1-second initializations. We use WOSAC to evaluate short time-horizon (8 seconds) simulation performance, and we further extend the evaluation to a longer time horizon (60 seconds) to systematically compare model-based and data-driven simulators. To support this effort, we developed \textit{Waymo2SUMO}, a robust toolchain for integrating WOMD scenarios into SUMO. This toolchain automates the conversion of high-definition maps and dynamic agent data into SUMO-compatible formats, including road network construction, traffic signal configuration, and agent behavior parameterization. This process is challenging due to the complexity of real-world traffic environments, encompassing diverse road users and intricate road geometries, as well as the significant disparity between SUMO’s road network representation and the vectorized map formats widely employed in the AV industry. By bridging these gaps, our methodology enables transportation researchers to exploit high-quality AV datasets in microscopic traffic simulators, facilitating new applications and insights. The toolchain is also adaptable to other datasets, such as NuScenes~\cite{caesar2020nuscenes-NuScenes} and NuPlan~\cite{caesar2021nuplan}, and can be extended to alternative model-based simulators.

Our findings reveal key trade-offs between model-based and data-driven simulation approaches. For short time-horizon simulations, SUMO achieves a meta metric (an overall metric considering kinematics, interaction, and map adherence) of 0.653 on the WOSAC benchmark. However, it is noteworthy that the SOTA data-driven method includes up to 10 million model parameters, compared to SUMO’s less than 100 parameters, making SUMO around 100,000x more lightweight. This underscores the trade-off between fidelity and computational efficiency. For long time-horizon simulations, data-driven methods suffer from distribution shift issues, with modeling errors compounding over time in autoregressive simulations. These errors lead to unrealistic behaviors, such as off-road driving, collisions, and incorrect lane usage. In contrast, SUMO demonstrates robustness over extended horizons, critical for large-scale AV development and validation. These results underscore the complementary strengths of model-based and data-driven simulators. Combining these paradigms could lead to high-fidelity, efficient, and robust traffic simulations.

This study seeks to bridge the gap in understanding the performance of model-based simulators in AV applications and to provide a foundation for future advancements in traffic simulation. The key contributions are as follows:
\begin{itemize}
    \item \textbf{Benchmarking SUMO with real-world datasets:} For the first time, we quantitatively evaluate the performance of SUMO using WOMD, providing an objective and transparent benchmark for model-based simulators.

    \item \textbf{Systematic comparison with data-driven simulators:} We provide a comparative analysis of SUMO and SOTA data-driven simulators across short and long-time horizons, offering insights into their strengths and limitations, and facilitating future developments.

    \item \textbf{Development of \textit{Waymo2SUMO} Toolchain:} We present a fully automated toolchain to integrate WOMD scenarios into SUMO, enabling the transportation research community to leverage cutting-edge AV datasets effectively. The code is open source for public use.

\end{itemize}
The remainder of this paper is organized as follows: Section~\ref{sec:relared_work} reviews related works on simulation development. Section~\ref{sec:methodology} describes the methodology for developing \textit{Waymo2SUMO} to integrate WOMD scenarios into SUMO. Section~\ref{sec:results} presents simulation results and comparative analyses for short and long time-horizons. Section~\ref{sec:discussion} discusses the limitations and advantages of model-based and data-driven simulators, providing directions for future research. Finally, Section~\ref{sec:conclusion} concludes the paper.

%% file: sections/related.tex
%%%%%%%%%%%%%%%%%%%%%%%%%%%%%%%%%%%%%%%%%%%%%%%%%%%%%%%%%%%%%%%%%%%%%%%%%%
\section{Related Work}
\label{sec:relared_work}

\subsection{Simulation platforms and scenario standards}
Traffic and AV simulation spans a spectrum of platforms.
Model-based microscopic traffic simulators such as SUMO~\cite{krajzewicz2012sumo}, VISSIM~\cite{fellendorf2010vissim}, and AIMSUN~\cite{casas2010aimsun} represent decades of development in transportation engineering, emphasizing structured road networks, parameterized behavior models, interpretability, and computational efficiency.
Complementing them, end-to-end AV simulators such as CARLA~\cite{dosovitskiy2017carla}, LGSVL~\cite{rong2020lgsvl}, and AirSim~\cite{shah2017airsim} enable testing full autonomy stacks with sensors and environment rendering.
To improve interoperability across tools and workflows, standardized scenario/map formats have been promoted in the AV ecosystem, and scenario-based testing frameworks (e.g., Paracosm) further support programmatic specification of interactive situations~\cite{majumdar2021paracosm}.

\subsection{Data-driven traffic simulation}
Learning-based simulators have rapidly progressed with large-scale trajectory datasets and modern generative/sequential modeling.
In academia, representative closed-loop sim-agent methods include TrafficSim~\cite{suo2021trafficsim}, BITS~\cite{xu2022bits}, Symphony~\cite{igl2022symphony}, TrafficBots~\cite{zhang2023trafficbots}, and SMART~\cite{wu2024smart}, while recent approaches explore diffusion-based controllable generation~\cite{jiang2023motiondiffuser,jiang2024scenediffuser} and transformer-style modeling~\cite{yuan2021agentformer}.
Industrial systems also report substantial investments in data-driven simulation stacks, such as Waymo SimulationCity~\cite{waymo_simulationcity}, NVIDIA's AV simulation toolchain~\cite{nvidia_av_simulation}, and Waabi World~\cite{waabi_world}.
Compared with model-based simulators, data-driven approaches can achieve high short-horizon fidelity but may encounter error accumulation and distribution shift in long-horizon autoregressive rollouts; this issue is closely related to classical compounding-error phenomena studied in imitation learning (e.g., DAgger)~\cite{ross2011dagger}.

\subsection{Benchmarks, datasets, and toolchains for closed-loop evaluation}
Public datasets and benchmarks are crucial for reproducible evaluation of interactive simulation.
WOMD provides large-scale real-world driving scenarios with map context and agent trajectories~\cite{ettinger2021large-womd}, and WOSAC standardizes closed-loop sim-agent evaluation with unified metrics and protocols~\cite{montali2024waymo-WOSAC-benchmark}.
Waymax further facilitates large-scale experimentation via an accelerated, data-driven simulator interface for batched evaluation and learning workflows~\cite{gulino2023waymax}.
Beyond the Waymo ecosystem, datasets and benchmarks such as nuScenes~\cite{caesar2020nuscenes-NuScenes} and nuPlan~\cite{caesar2021nuplan} enrich the landscape for perception/forecasting/planning research, while research platforms like Nocturne~\cite{vinitsky2022nocturne} and MetaDrive~\cite{li2021metadrive} support scalable multi-agent driving experimentation.
Meanwhile, bridging heterogeneous HD-map/scenario representations and simulation engines remains a practical challenge.
Lane-level HD map abstractions (e.g., Lanelet2)~\cite{poggenhans2018lanelet2} and conversion toolchains such as the CommonRoad Scenario Designer~\cite{maierhofer2021commonroad} and automatic OpenSCENARIO-to-CommonRoad conversion~\cite{lin2023openscenario2commonroad} highlight ongoing efforts toward interoperability.
Our \textit{Waymo2SUMO} complements these works by providing an automated bridge from WOMD/WOSAC scenarios to SUMO, enabling systematic evaluation of a widely used model-based simulator on large-scale real-world AV datasets.

%% file: sections/method.tex
%%%%%%%%%%%%%%%%%%%%%%%%%%%%%%%%%%%%%%%%%%%%%%%%%%%%%%%%%%%%%%%%%%%%%%%%%%
\section{Methodology}
\label{sec:methodology}

\input{sections/method_framework}
\input{sections/method_static}

\input{sections/method_dynamic}

%% file: sections/method_framework.tex
\subsection{Overall Framework}

The overall framework for converting a real-world scenario in the WOMD into the SUMO traffic simulator is shown in Figure \ref{fig:framework}. Given a raw WOMD scenario that contains static map features, traffic signals, and historical agent trajectory data, the integration pipeline involves two key steps: \textit{static context setup} and \textit{dynamic agent setup}. These steps correspond directly to the required inputs for the SUMO simulation environment.

The static context setup includes converting the map data and estimating traffic signals at signalized intersections, as detailed in Section \ref{sec:static-context}. Although the WOMD provides rich static context through high-precision vectorized maps with semantic attributes, these representations differ substantially from the formats needed by traffic simulators like SUMO. For example, in the WOMD, map features such as lane centers, road edges, road lines, and pedestrian crossings are represented as continuous vector segments, which include additional semantics such as neighborhood relations, road boundaries, and right-of-way information. In contrast, SUMO represents road networks using nodes, edges, and connections. The data structure for traffic signals is also distinct between the WOMD and SUMO. Therefore, an essential prerequisite is establishing a comprehensive toolchain to convert real-world scenarios into SUMO-compatible road networks.

The dynamic agent setup involves defining simulation parameters, initializing agents with corresponding parameters, and applying extra control logics within the simulator. These steps  will be further explained in Section \ref{sec:dynamic-agents}. In SUMO, simulation behaviors are influenced by a range of tunable parameters, some of which apply to the entire scenario, while others govern individual agent behaviors. In our pipeline, these parameters are determined at the start of the simulation based on the topologies of the converted road network and the historical trajectories of the agents. Once all parameters are set, agents are initialized into the simulator. Additionally, extra control logics are implemented to account for specific agent behaviors, compensating for some of SUMO’s inherent limitations as a traffic simulator.

Together, these two modules provide the necessary road network and traffic demand inputs for SUMO. SUMO then generates 32 stochastic 8-second, 10Hz motion sequences for each scenario, which are ready for evaluation in the Waymo Open Sim Agents Challenge (WOSAC), as shown in Figure \ref{fig:framework}(c).

%% file: sections/method_static.tex
\subsection{Scenario Static Context Setup}
\label{sec:static-context}
\subsubsection{Road Network Construction}
\label{sec:map-conversion}

\begin{figure*}[t]
    \centering
    \includegraphics[width=0.9\linewidth]{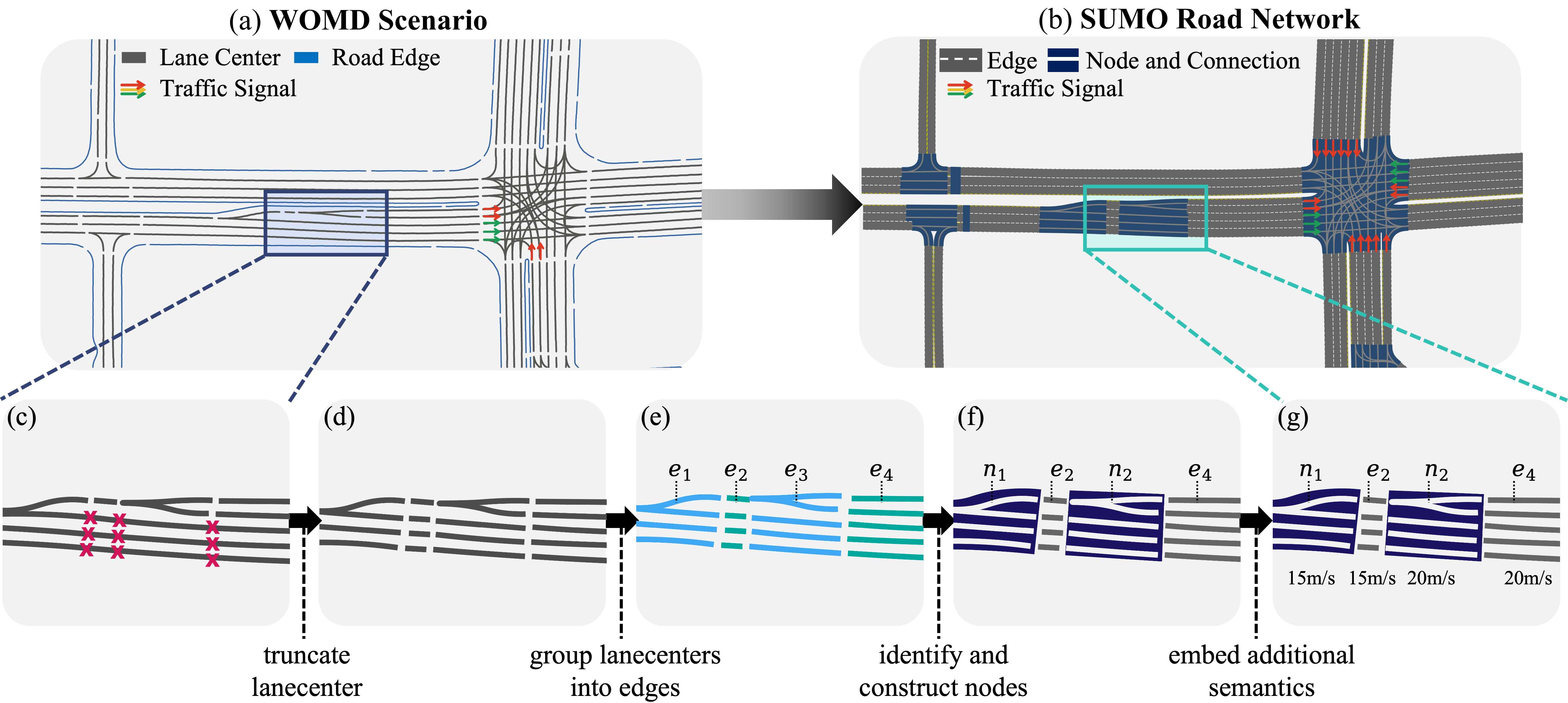}
    \caption{The road network construction process.}
    \label{fig:map-conversion}
\end{figure*}

A real-world scenario in the WOMD\footnote{For a detailed definition of road map in the WOMD, please refer to \url{https://waymo.com/open/data/motion/}} represents road geometry using lane centers, denoted as $\mathcal{L} = \{l_i\}, i=1,\dots, |\mathcal{L}|$, where each lane center is described by a continuous sequence of points. These lane centers carry information about lane types, neighboring lanes, boundaries, and entry/exit points. Figure \ref{fig:map-conversion}(a) illustrates this representation, where each grey line represent a lane center segment $l_i$. Road edges are depicted by narrow blue lines. Traffic signals are also included, represented as signal heads associated with specific lane centers.

\begin{figure*}[t]
    \centering
    \includegraphics[width=0.8\linewidth]{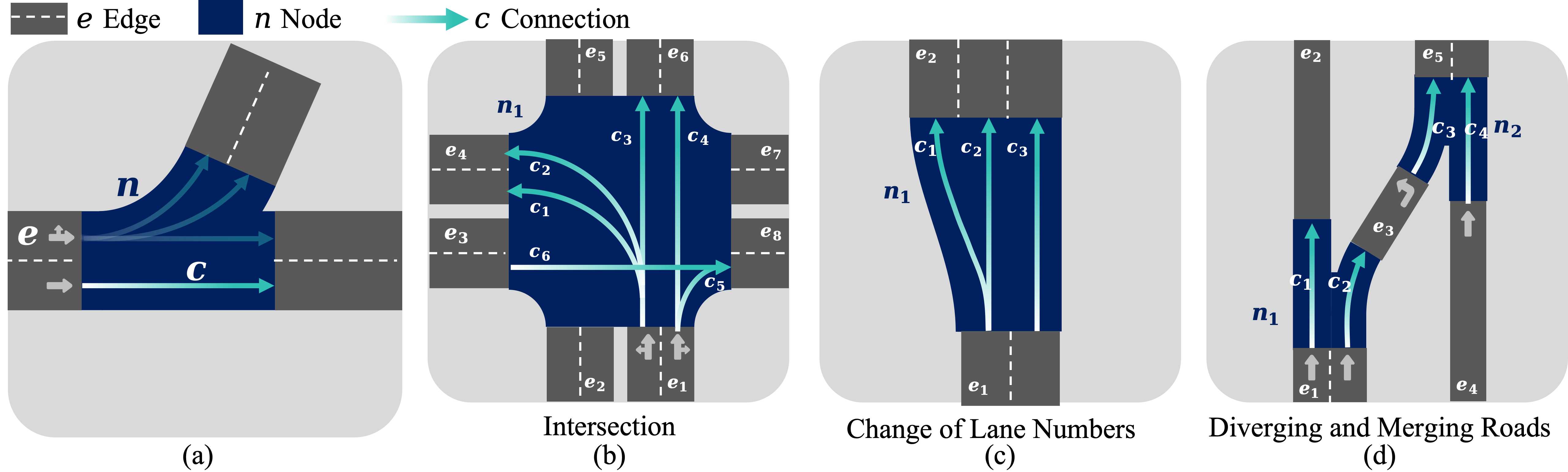}
    \caption{Definition of nodes, edges, and connections in SUMO. (a) The general structure around a node, (b) A node representing an intersection, (c) A node representing an area with changing number of lanes, (d) A node representing an area of diverging or merging roads.}
    \label{fig:sumo-definition}
\end{figure*}

In contrast, a SUMO road network \footnote{For a more detailed definition of SUMO’s road network structure, please refer to \url{https://sumo.dlr.de/docs/Networks/SUMO\_Road\_Networks.html}} (Figure \ref{fig:map-conversion}(b)) is structured as a directed graph comprising three sets of elements, edges $\mathcal{E}$, nodes $\mathcal{N}$ (also known as junctions), and connections $\mathcal{C}$. These elements are further illustrated in Figure \ref{fig:sumo-definition}(a):

\begin{itemize}
    \item \textbf{Edges}, denoted as $\mathcal{E} = \{e_i\}, i=1,\dots,|\mathcal{E}|$, represent unidirectional road segments consisting of multiple parallel lanes. Each edge starts at one node and ends at another, with associated attributes such as speed limits, lane width, and priority information.
    \item \textbf{Nodes}, denoted as $\mathcal{N}=\{n_i\}, i=1,\dots, |\mathcal{N}|$, represent critical areas where traffic flows intersect, diverge, or merge. Nodes can be intersections (Figure \ref{fig:sumo-definition}(b)), segments with changing lane counts (Figure \ref{fig:sumo-definition}(c)), or areas where roads split or merge (Figure \ref{fig:sumo-definition}(d)). Each node includes right-of-way rules and traffic signal programs, where applicable.
    \item \textbf{Connections}, denoted as $\mathcal{C}=\{c_i\}, i=1,\dots, |\mathcal{C}|$, link individual lanes between nodes, specifying valid movement directions between connected lanes. For example, in Figure \ref{fig:sumo-definition}(c), $c_1$, $c_2$, and $c_3$ are three connections within node $n_1$, each connecting a lane in $e_1$ to a lane in $e_2$.
\end{itemize}

The challenge in integrating real-world map data from the WOMD into SUMO lies in the different ways that the two systems represent map semantics. Therefore, we propose the following conversion method, which maps lane centers \( \mathcal{L} \) from the WOMD into the SUMO network, described by \( \mathcal{E} \) (edges), \( \mathcal{N} \) (nodes), and \( \mathcal{C} \) (connections).

\textbf{Truncating Lane Centers.} In SUMO, all lanes within an edge are parallel and typically share similar lengths, while it is not the case for lane centers in the WOMD. However, the WOMD stores lane adjacency information explicitly for each pair of adjacent lane centers \( \{\ell_i, \ell_j\} \), including the start and end points of adjacency. For each lane center \( \ell_i \in \mathcal{L} \), we evaluate the adjacency relationship with every neighboring lane center \( \ell_j \). If either the adjacency start or end point between \( \ell_i \) and \( \ell_j \) does not coincide with the corresponding start or end point of \( \ell_i \), the lane center \( \ell_i \) is split at that point, creating two new lane centers, denoted \( \ell_{i1} \) and \( \ell_{i2} \). This truncation process continues recursively until no further splits are possible, resulting in a refined set of lane centers \( \mathcal{L}^{\prime} \). The completion of this step ensures that every pair of neighboring lane centers shares the same length, which forms the basis for the subsequent grouping into edges.

To illustrate this process, Figure \ref{fig:map-conversion}(c) shows a segment from the example scenario in Figure \ref{fig:map-conversion}(a). The lane center labeled as \( \ell_1 \), highlighted in orange, is recursively split three times. This is because each of its adjacent left neighbors—\( \ell_2 \), \( \ell_3 \), \( \ell_4 \), and \( \ell_5 \)—has either a starting or ending point that does not align with the corresponding endpoints of \( \ell_1 \). As a result, \( \ell_1 \) is divided into four segments, labeled as \( \ell_6 \), \( \ell_7 \), \( \ell_8 \), and \( \ell_9 \). This splitting process propagates further: the common right neighbor of \( \ell_6 \), \( \ell_7 \), \( \ell_8 \), and \( \ell_9 \) must also be split into matching parts to ensure alignment between connected lane centers. As shown in Figure \ref{fig:map-conversion}(d), the recursive splitting continues until all lane centers are segmented such that each adjacent pair shares nearly identical start and end points.

\textbf{Grouping Lane Centers into Edges.} After truncation, the refined lane center set \( \mathcal{L}^{\prime} \) is divided into disjoint subsets, each subset representing an edge \( e_k \) in the SUMO network. We denote these disjoint subsets as
$$
\mathcal{L'} = \biguplus_{k} S_e^k, \quad S_e^k \subseteq \mathcal{L'}, \quad S_e^i \cap S_e^j = \emptyset \text{ for } i \neq j
$$
where each \( S_k \) is a disjoint subset containing parallel lane centers that belong to the same edge \( e_k \). Denote the set of all such disjoint subsets as $\mathcal{S}_e = \{ S_e^k \}$. Mathematically, each edge \( e_k \) is defined as:
$$
e_k := S_e^k = \{ \ell_{k1}', \ell_{k2}', \dots, \ell_{km}' \}, \quad e_k \in \mathcal{E}
$$

Visually, this grouping process transforms the 18 truncated lane centers from Figure \ref{fig:map-conversion}(d) into four distinct edges—\( e_1 \), \( e_2 \), \( e_3 \), and \( e_4 \)—as shown in Figure \ref{fig:map-conversion}(e). Each edge \( e_k \) consists of a subset of parallel lane centers aligned along the same road segment.

\textbf{Identifying and Constructing Nodes.} Unlike SUMO, the road map in the WOMD does not explicitly define nodes. To construct the set of nodes \( \mathcal{N} \), we use a heuristic based on two geometric patterns commonly observed in node structures in SUMO road networks:
\begin{itemize}
    \item Diverging connections: Within a node, some pairs of connections diverge from nearly identical start points and end in different directions (e.g., \( c_2 \) and \( c_3 \) in Figure \ref{fig:sumo-definition}(b), $c_1$ and $c_2$ in Figure \ref{fig:sumo-definition}(c)).
    \item Merging connections: Some other pairs of connections merge from distinct starting points to nearly identical end points (e.g., \( c_3 \) and \( c_4 \) in Figure \ref{fig:sumo-definition}(d)).
\end{itemize}

This implies that all connections within the same node can be identified and grouped using a union-find algorithm. Take the intersection node \( n_1 \) in Figure \ref{fig:sumo-definition}(b) as an example. Connections \( c_1 \) and \( c_2 \) can be grouped together because they diverge from a nearly identical start point and head in different directions, satisfying the diverging connection heuristic. Similarly, the pairs \( c_2 \) and \( c_3 \), as well as \( c_4 \) and \( c_5 \), are identified to belong to the same node based on the same heuristic. Additionally, since \( c_3 \) and \( c_4 \) are parallel neighbors, they are also grouped into the same node. Through this iterative process, we infer that the set of connections \( \{c_1, c_2, c_3, c_4, c_5\} \) belong to the same node \( n_1 \). This grouping captures all connections diverging from the same incoming approach. Furthermore, \( c_5 \) and \( c_6 \) are identified as merging connections, as they converge into a nearly identical endpoint. This satisfies the merging connection heuristic, confirming that they both belong to the same node \( n_1 \). Repeating this process for all remaining connections, every connection within this intersection will be grouped together and assigned to node \( n_1 \).

Therefore, using this heuristic, nodes \( \mathcal{N} \) are constructed by grouping \textit{some of} the edges in \( \mathcal{E} \) into a series of edge subsets, denoted as $\mathcal{S}_n = \{ S_n^k \}, S_n^k \in \mathcal{E}$. Specifically, 
\begin{itemize}
    \item For any edge \( e_i \in \mathcal{E}\), if there exists lane centers \( \ell_m, \ell_n \in e_i \) that satisfy one of the two geometric patterns above, then \( e_i\) is an element of a subset.
    \item For any two edges \( e_j, e_k \in \mathcal{E} \), if there exists lane centers \( \ell_p \in e_j \) and \( \ell_q \in e_k \) that satisfy one of the two geometric patterns above, then \( e_j \) and \( e_k \) are grouped into the same subset.
\end{itemize}

Once the entire edge set \( \mathcal{E} \) has been processed in this manner, we obtain disjoint subsets of edges, where each subset is represented as
$$
S_n^k = \{ e_{v1}, e_{v2}, \dots, e_{vk} \} \subseteq \mathcal{E}
$$

Not that some edges do not belong to any subset. For each subset \( S_n^k \), a node \( n \in \mathcal{N} \) is instantiated. The shape of the node is determined by the union of the shapes of all edges \( e_i \in S_n^k \):
$$
\text{Shape}(n) = \bigcup_{e \in S_n^k} \text{Shape}(e)
$$
Additionally, each lane center \( \ell_i \in e_j, \, e_j \in S_n \), is converted into a connection \( c \in \mathcal{C} \). Once an edge has been grouped into a node subset, it is removed from the edge set \( \mathcal{E} \). After processing all edge subsets \( S_n^k \in \mathcal{S}_n \), we obtain the final node set \( \mathcal{N} \), the updated edge set \( \mathcal{E}^{\prime} \), and the connection set \( \mathcal{C} \), finalizing the road network topology.

The above process is visually illustrated in Figure \ref{fig:map-conversion}(f). Edge $e_1$ itself forms a subset $S_n^1 = \{ e_1\}$, because lanes within $e_1$ exhibit the diverging geometric pattern. This subset is transformed into node $n_1$. Similarly, edge $e_3$ itself forms a subset $S_n^2 = \{ e_3\}$ for the same reason, and it is transformed into node $n_2$. These two edges are no longer elements of the edge set $\mathcal{E}$.

\textbf{Embedding Additional Semantics.} Lastly, additional information from the WOMD is transferred to SUMO where compatible, as in Figure \ref{fig:map-conversion}(g). This includes:
\begin{itemize}
    \item For an edge $e$, its speed limits and lane width;
    \item For a node $n$, which movements are regulated by a stop sign (if the node represents an un-signalized intersection), and the traffic signal state for each movement (if the node represents a signalized intersection).
\end{itemize}

This completes the map conversion. The pseudo-code for this conversion process is provided in Algorithm \ref{alg:map-conversion}.

\begin{algorithm}[h!]
\caption{Convert WOMD Lane Centers to SUMO Road Network}
\begin{algorithmic}[1]
\Require Lane center set \( \mathcal{L} \) from WOMD
\Ensure SUMO network with edges \( \mathcal{E} \), nodes \( \mathcal{N} \), and connections \( \mathcal{C} \)
\State Initialize empty sets \( \mathcal{E}, \mathcal{N}, \mathcal{C}, \mathcal{L'} \)

\While{there exists \( \ell_i \in \mathcal{L} \) that can be split}
    \State Find the split point between \( \ell_i \) and a neighbor \( \ell_j \)
    \State Split \( \ell_i \) into \( \ell_{i1} \) and \( \ell_{i2} \)
    \State \( \mathcal{L'} \gets \mathcal{L'} \cup \{ \ell_{i1}, \ell_{i2} \} \)
\EndWhile

\State Partition \( \mathcal{L'} \) into disjoint subsets $
\mathcal{L'} = \biguplus_{k} S_e^k, \quad S_e^k \subseteq \mathcal{L'}, \quad S_e^i \cap S_e^j = \emptyset \text{ for } i \neq j$

\For{each subset \( S_e^k \in \mathcal{S}_e \)}
    \State Create edge \( e_k := S_k \)
    \State \( \mathcal{E} \gets \mathcal{E} \cup \{ e_k \} \)
\EndFor

\For{each edge \( e_i \in \mathcal{E} \)}
    \If{some $\ell_m, \ell_n \in e_i$ satisfy heuristics}
        \State Add $e_i$ to a new subset $S_n^k$
    \EndIf
    \For{each edge \( e_j \in \mathcal{E} \setminus \{ e_i \} \)}
        \If{some \( \ell_p \in e_i \) and \( \ell_q \in e_j \) satisfy heuristics}
            \State Group \( e_i \) and \( e_j \) into a common subset \( S_n^k \)
        \EndIf
    \EndFor
\EndFor

\For{each subset \( S_n^k \in \mathcal{S}_n \)}
    \State Create node \( n \) representing \( S_n^k \)
    \State \( \mathcal{N} \gets \mathcal{N} \cup \{ n \} \)
    \For{each \( \ell_i \in e_j, \, e_j \in S_n \)}
        \State Create connection \( c \)
        \State \( \mathcal{C} \gets \mathcal{C} \cup \{ c \} \)
    \EndFor
    \State \( \mathcal{E} \gets \mathcal{E} \setminus S_n \)
\EndFor

\State \Return \( \mathcal{E}, \mathcal{N}, \mathcal{C} \)

\end{algorithmic}
\label{alg:map-conversion}
\end{algorithm}

\subsubsection{Traffic Signal Estimation}

For signalized intersections, complete traffic signal states are required in SUMO to perform simulations. While the traffic signal semantics in raw data are indeed transferred to SUMO road networks, the signal state records are almost always incomplete. This issue is a result from the mechanism the traffic signal data is collected: AV sensors capture the traffic lights that are applied to their approach only, leading to missing data for other approaches. Furthermore, computer vision techniques for camera data processing can introduce errors, resulting in inaccurate traffic signal states.

These two deficiencies in the raw data necessitate imputation and rectification of the traffic signal data. We estimate the traffic signal states by leveraging the available raw signal data, historical vehicle trajectories, and transportation domain knowledge. We first identify signalized intersections in each scenario through geometric patterns, similar to the node identification process described in Section \ref{sec:map-conversion}. Once identified, the estimation process involves integrating the raw signal state, vehicle trajectory data, and the ring-barrier diagram, following NEMA standard commonly used in traffic signal control~\cite{urbanik2015signal}. Specifically, vehicle behaviors, such as acceleration and deceleration patterns near intersections, are analyzed to infer probable signal states. For instance, vehicles approaching a stop bar at high speed without slowing down are likely reacting to a green signal. Furthermore, missing or erroneous states are corrected using these behavioral indicators along with predefined signal patterns constrained by NEMA. For a detailed explanation of this method, please refer here~\cite{yan2026improving}. As only the first second of traffic signal states and vehicle trajectories are provided in the WOMD, we extend the signal states beyond the historical data by holding the last known state constant for the remainder of the simulation window.

With the converted road networks and traffic signal programs, the static context setup for SUMO simulation is complete.

%% file: sections/method_dynamic.tex
\subsection{Scenario Dynamic Agents Setup}
\label{sec:dynamic-agents}
\subsubsection{Agent Parameters Setup}

\begin{table*}[t]
    \centering
    \caption{Agent parameters setup.}
    \footnotesize
    
    \begin{tabular}{p{1cm}cp{6cm}c}
        \toprule
        \textbf{Model} & \textbf{Parameters} & \textbf{Description} & \textbf{Value} \\
        \midrule
        \multirow{6}{=}{Car-following model} 
        & speedFactor & The vehicles expected multiplier for speed limits & $N(\max(0.75, v_{\text{history}} / v_{\text{limit}}), 0.1)$ \\
        & minGap & Minimum Gap when standing (m) & $\text{Truncnormal}(2.5, 0.5^2; 0, 5)$ \\
        & accel & The acceleration ability of the vehicles & $\text{Truncnormal}(2, 0.2^2; 1, 4.5)$ \\
        & decel & The deceleration ability of the vehicle & $\text{Truncnormal}(2.5, 0.2^2; 1, 4.5)$ \\
        & sigma & Driver imperfection factor & $\text{Truncnormal}(0.5, 0.2^2; 0, 1)$ \\
        & tau & Driver's reaction time & $\text{TruncLognormal}(0, 0.1^2; 0, 5)$ \\
        & startupDelay & The delay time before starting to drive after having had to stop & $\text{Exponential}(3)$ \\
        \midrule
        \multirow{3}{=}{Lane-changing model} 
        & minGapLat & The desired minimum lateral gap to other vehicles (m) & $\text{Truncnormal}(0.6, 0.08^2; 0.4, 0.8)$ \\
        & lcKeepRight & The eagerness for following the obligation to keep right & $\text{TruncLognormal}(100, 0.1^2; 0, 1.5)$ \\
        & lcSublane & Tendency to use sublanes & $\text{Truncnormal}(0.4, 0.3^2; 0, 10)$ \\
        \midrule
        \multirow{3}{=}{Intersection model} 
        & jmStopLineGap & Stopping distance in front of a stop line & $\text{Lognormal}(0.4, 0.5^2) + 1$ \\
        & jmSigmaMinor & Driving imperfection while passing a minor link & $\text{Truncnormal}(0.5, 0.2^2; 0, 1)$ \\
        & jmIgnoreKeepClearTime & The maximum of waiting time before entering an intersection & \\
        \bottomrule
    \end{tabular}
    \label{tab:parameters}
\end{table*}

SUMO's microscopic simulation revolves around three key models: the car-following model, the lane-changing model, and the intersection model.

\begin{itemize}
    \item \textbf{Car-Following Model}: This model simulates how individual vehicles adjust their speed and distance in response to the vehicle in the front, capturing realistic behaviors like acceleration, deceleration, and stopping. While SUMO implements various car-following models, the default Krauß model is adopted in this study \cite{krauss1997-sumo-carfollowing}.
    \item \textbf{Lane-Changing Model}: This model focuses on the decision-making process of when and how vehicles switch lanes. While SUMO's default lane-changing model is LC2013 \cite{erdmann2015-sumo-lanechange}, which simulates instantaneous lane changes, we opt for SL2015 to enable sublane simulation for greater realism. This allows for smoother, more gradual lane transitions.
    \item \textbf{Intersection Model}: The intersection model manages how vehicles interact at intersections, integrating traffic light control and right-of-way behaviors to ensure safe and smooth navigation through crossroads \cite{ermann2014sumo-intersection}.
\end{itemize}

Each of these models allows for adjustable parameters on a per-agent basis, providing flexibility in simulating a wide range of driving behaviors. In our study, some parameters, such as the speed factor, are customized using data from historical trajectories. Others, like driving imperfections and lane-changing eagerness, are randomized to introduce variability and stochasticity in the simulation. We detail how we configure these parameters in Table \ref{tab:parameters}.

In addition to the model-specific parameters, each agent in SUMO requires a predefined route, which is not explicitly provided in the WOMD raw data. We heuristically determine these routes by applying depth-first search (DFS) algorithm. During DFS, two additional considerations are made. First, when selecting the next road segment, main roads are given a significantly higher probability of being chosen over side roads. Second, the road segment is also chosen in a way to avoid excessive lane changes within short distances.

Setting up all these agent parameters formulates the traffic demand for a scenario. %A more comprehensive list of parameters setup can be found in Appendix \ref{sec:appendix_parameters}.

\subsubsection{Extra Control Logics}
While SUMO excels at simulating traffic flow, it has inherent limitations when applied to real-world scenarios, further discussed in Section \ref{sec:discussion-limitations}. First, SUMO predominantly focuses on traffic-related behaviors of agents, which are typically confined to road networks. However, real-world road users often exhibit more nuanced behaviors that SUMO is not equipped to simulate. These include, but are not limited to, on-street parking, maneuvering into parking lots, reversing, and occasionally crossing over the stop line when waiting for a red signal. Second, due to the limitations in the static map conversion process and the inherent constraints of the SUMO network format, certain imperfections arise in the constructed road networks. These imperfections make it difficult to place all agents onto the road network and fully automate motion simulation. For example, vehicles located in driveways cannot be properly placed on SUMO’s road network, as driveways are not represented in the converted maps.

\begin{figure}[h!]
    \centering
    \includegraphics[width=1\linewidth]{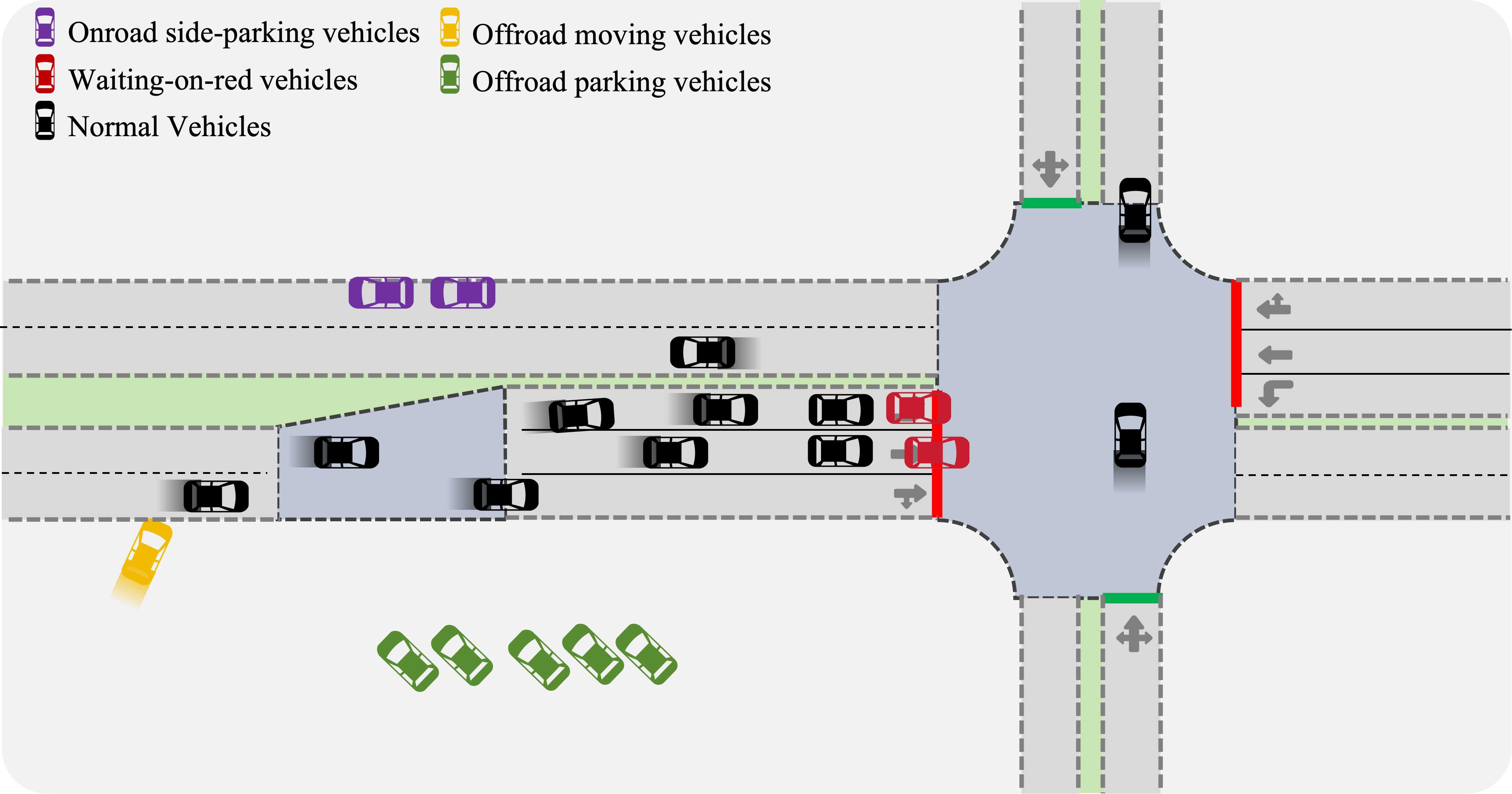}
    \caption{Illustration of extra control logics applied to agents with special behaviors.}
    \label{fig:extra-control}
\end{figure}

To address these shortcomings, we implemented additional control logics to better accommodate various agent behaviors commonly seen in real-world scenarios:

\begin{itemize}
    \item \textbf{Vehicles waiting at red signals}:  
    In the real world, vehicles sometimes edge slightly past the stop line while waiting at a red light (depicted by the red vehicles in Figure \ref{fig:extra-control}). However, SUMO does not account for this behavior and would immediately accelerate the vehicle into the intersection if no additional control is applied.  
    
    To address this, a command is issued to keep the agent stationary if it is located on the road network, is within a distance $d_{\text{intersection}}$ from the stop line of a signalized intersection, has a red signal, and its historical speed has been zero. An exception is made for vehicles in the rightmost lane, as right turns are typically permitted on red lights.

    \item \textbf{Parked vehicles on the roadside}:  
    SUMO cannot simulate vehicles parking on the roadside (purple vehicles in Figure \ref{fig:extra-control}) unless a designated parking area is explicitly defined in the road network. Without additional control, such a vehicle would begin moving immediately upon instantiation.

    To resolve this, a command is issued to keep the vehicle stationary if it is on the road network and has maintained a speed of zero. Additionally, the vehicle remains stationary if its distance to the nearest road edge is below the threshold $d_{\text{roadedge}}$, or if its distance to the nearest lane center exceeds the threshold $d_{\text{lanecenter},1}$.

    \item \textbf{Vehicles off the road network}:  
    SUMO lacks the capability to simulate vehicles that operate outside the road network. This creates challenges for simulating agents located off-road, such as in parking lots or other non-road areas, as depicted by the green and yellow vehicles in Figure \ref{fig:extra-control}.  

    To handle this, if an agent cannot be placed on the road network and its distance to the nearest lane center exceeds the threshold $d_{\text{lanecenter},2}$, the agent is kept stationary (green vehicles). If the agent is within the $d_{\text{lanecenter},2}$ threshold, it moves with a constant linear and angular velocity matching its last recorded historical state (yellow vehicle).
\end{itemize}

These additional control logics enhance SUMO’s ability to simulate a broader range of agent behaviors and address scenarios where default SUMO behavior would fail to replicate real-world actions. %Implementation details can be found in Appendix \ref{sec:appendix_parameters}.

%% file: sections/results.tex
%%%%%%%%%%%%%%%%%%%%%%%%%%%%%%%%%%%%%%%%%%%%%%%%%%%%%%%%%%%%%%%%%%%%%%%%%%
\section{Results}
\label{sec:results}

To validate the effectiveness of our proposed toolchain for converting WOMD real-world scenarios into SUMO and systematically assess SUMO’s performance, we applied our method to the entire WOMD dataset, conducted extensive simulations, and evaluated the simulation rollouts on WOSAC. To the best of our knowledge, this work not only introduces the first fully automated toolchain for converting real-world scenarios into traffic simulators but also represents the first systematic benchmarking of a model-based simulator on large-scale datasets.

This section is organized as follows: Section \ref{sec:results-demonstration-static-context} demonstrates the adaptability of our static context setup process by presenting the converted road networks for a series of diverse scenarios. Section \ref{sec:results-scenario-analysis} analyzes a few selected scenarios, highlighting the realism of the simulation results and SUMO's ability to generate stochastic and diverse agent behaviors within the same scenario. Section \ref{sec:results-short-time-performance} provides quantitative metrics on WOSAC to evaluate SUMO’s performance in short time-horizon simulations. Finally, Section \ref{sec:results-long-time-performance} assesses SUMO’s performance in long time-horizon simulations, offering a more objective comparison with learning-based models.

\input{sections/results_static}
\input{sections/results_scenario}

\input{sections/results_wosac}

\input{sections/results_long}

%% file: sections/results_static.tex
\subsection{Demonstration of Static Context Setup Results}
\label{sec:results-demonstration-static-context}

\begin{figure}[ht]
    \centering
    \includegraphics[width=1\linewidth]{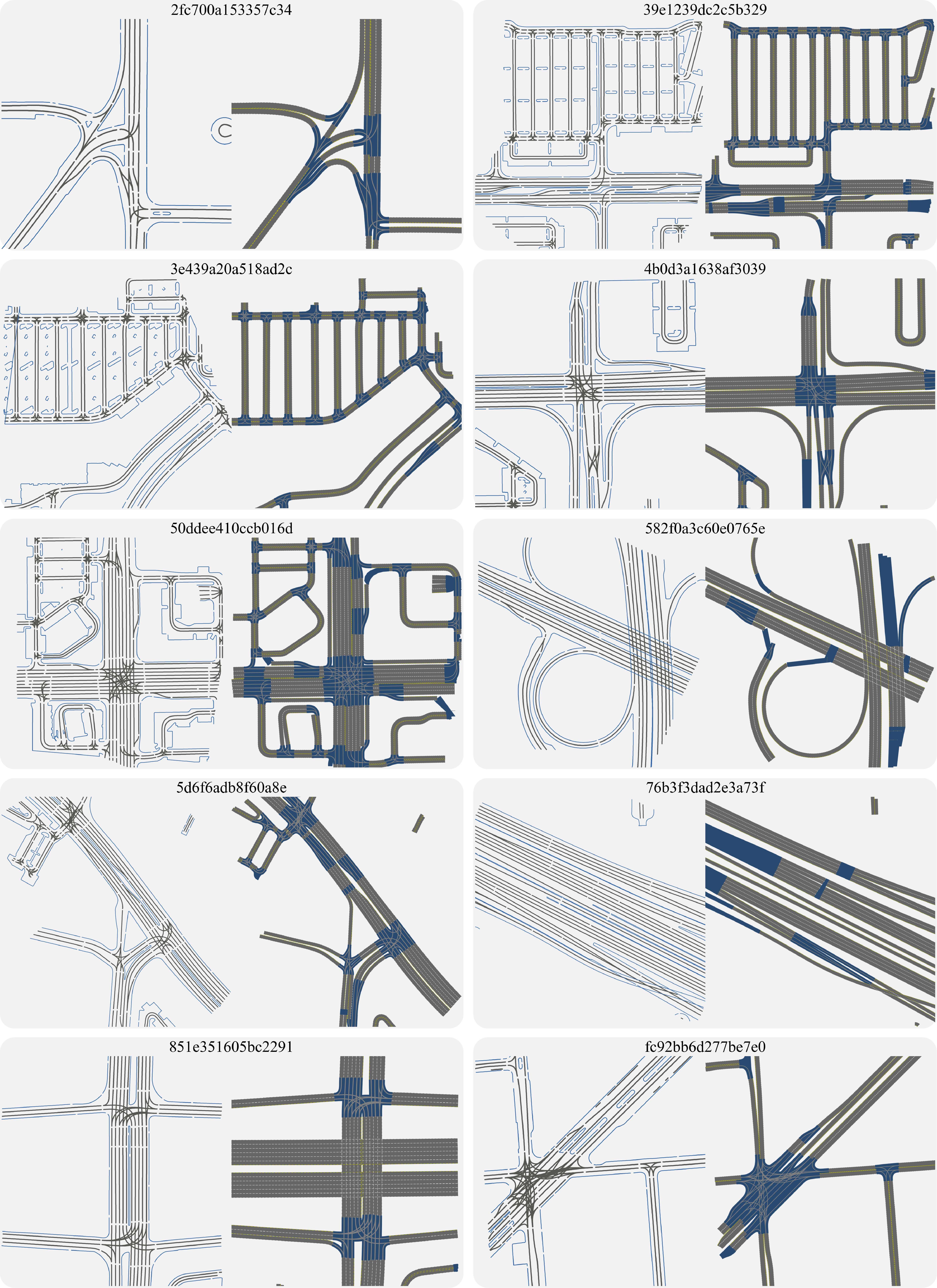}
    \caption{Demonstration of the static road network construction method in diverse real-world scenarios. The title for each pair of sub-figures is the corresponding scenario ID in the WOMD.}
    \label{fig:map-demonstration}
\end{figure}

Our static context conversion tool demonstrates broad adaptability, effectively handling a wide range of scenarios—from highways, intersections, and roundabouts to parking lots and even more irregular road structures. Figure \ref{fig:map-demonstration} illustrates the map conversion results for several scenarios from the WOMD. For each scenario, the left plot shows the original WOMD data, where lane centers and road edges are visualized; the right plot displays the corresponding SUMO road network generated through our our process.

For example, scenario 5d6f6adb8f60a8e represents a unique 3-way intersection. In this layout, a dedicated right-turn lane is separated from the main intersection by a large refuge island. Furthermore, this right-turn lane immediately intersects with a side road after bypassing the intersection. Despite these irregularities, our conversion method successfully maps this geometry into SUMO's network, with a combination of edges (in grey) and nodes (in dark blue).

Scenario fc92bb6d277be7e0 showcases an even more complex case: an irregular 6-way intersection supporting numerous turning directions. Our tool seamlessly consolidates all lane centers within the intersection into a single large node, and accurately represents the intersection's full set of possible movements.

%% file: sections/results_scenario.tex
\subsection{Scenario Analysis}
\label{sec:results-scenario-analysis}

\subsubsection{Scenario 1: Simulating Vehicles in a Signalized Four-Way Intersection}

\begin{figure}[ht]
    \centering
    \includegraphics[width=1\linewidth]{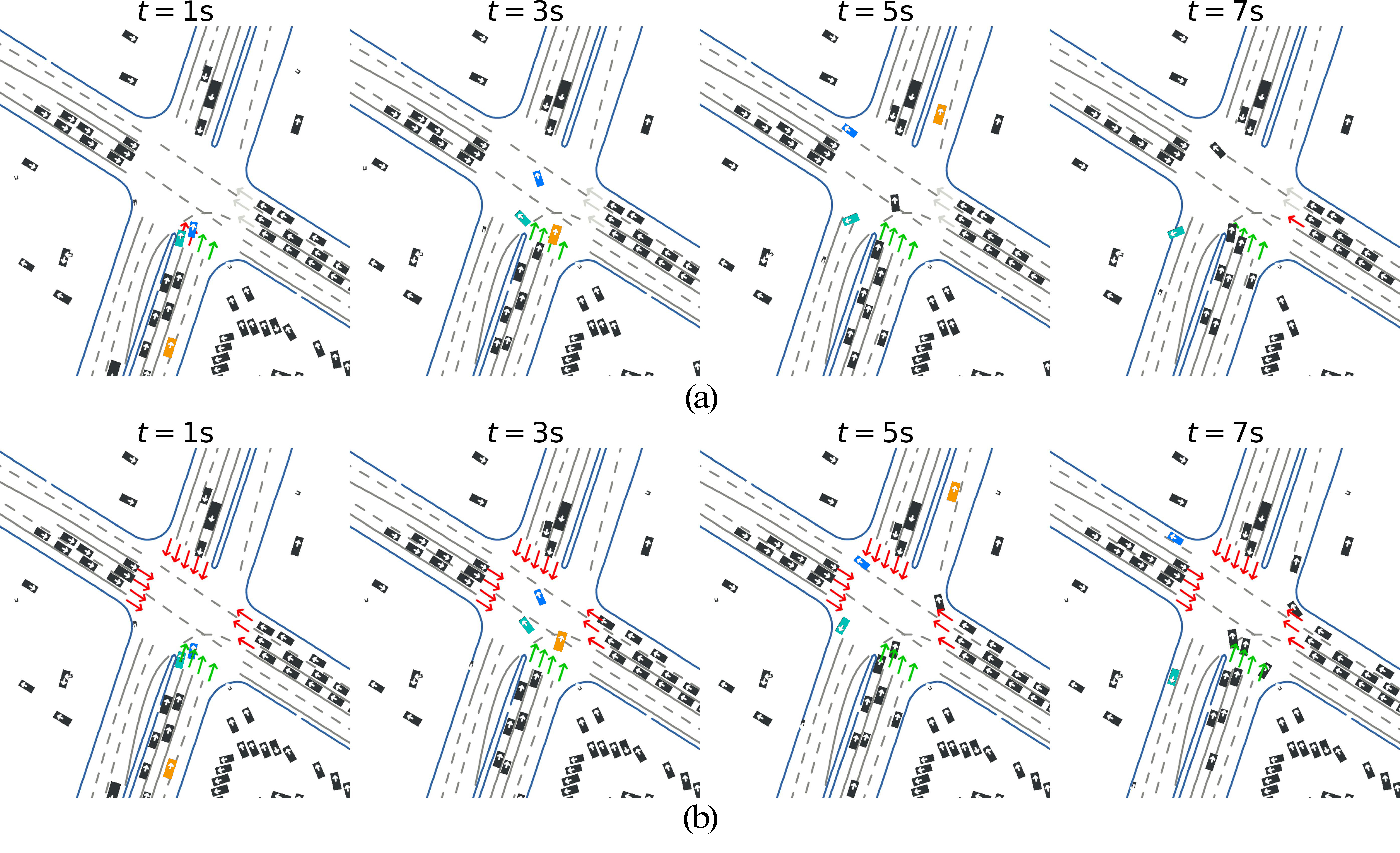}
    \caption{Demonstration of scenario 1. (a) Ground-truth, (b) Simulation rollouts.}
       \label{fig:scenario1}
\end{figure}

This scenario (ID: 2d84b1ab55ab81d3) focuses on a large signalized four-way intersection with high traffic volumes, depicted in Figure \ref{fig:scenario1}. All approaches to the intersection allow movements in every direction, with some permitting U-turns. The raw data—comprising the map, ground-truth vehicle trajectories, and incomplete traffic signals—are shown in Figure \ref{fig:scenario1}(a). For clarity, we present four moments within the 9-second data segment.

Figure \ref{fig:scenario1}(b) shows the simulation rollout generated by SUMO. As depicted, the traffic signals for the 9-second horizon are accurately estimated, correcting errors in the raw data. Notably, the signal state for the southbound left-turn movement is incorrectly recorded as red in the first two seconds. However, since the vehicles were observed moving in the raw data, this signal state was rectified during the context setup.

The simulated vehicle trajectories closely resemble the ground truth. Vehicles follow traffic rules according to the signals, and in the green-light southbound lane, two vehicles (highlighted in cyan) successfully perform U-turns from the leftmost lane, accurately replicated in the SUMO simulation.

\subsubsection{Scenario 2: Simulating Complex Vehicle Behaviors}

\begin{figure}[ht]
    \centering
    \includegraphics[width=1\linewidth]{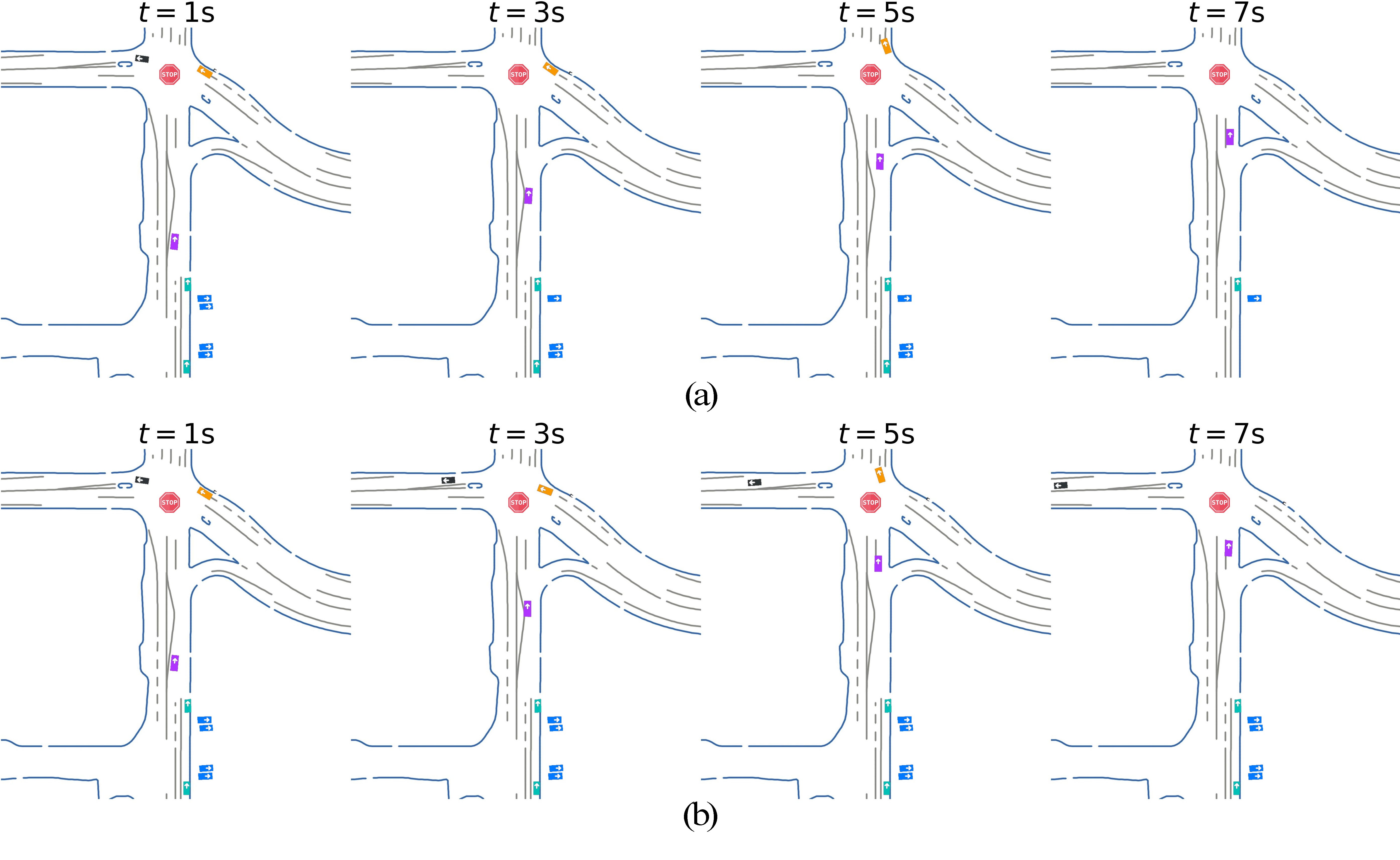}
    \caption{Demonstration of scenario 2. (a) Ground-truth, (b) Simulation rollouts.}
    \label{fig:scenario2}
\end{figure}

The second scenario (ID: 6b48214f47be32fc) features an all-way stop intersection, where all vehicles are required to come to a complete stop before proceeding. This rule is represented by the brown-colored lane centers within the intersection, as shown in Figure \ref{fig:scenario2}(a). Additionally, the scenario includes parked vehicles at the bottom: two roadside vehicles (boxed in red) and several off-road vehicles (boxed in purple).

These behaviors are reproduced in the simulation, as shown in Figure \ref{fig:scenario2}(b). At the intersection, an eastbound vehicle (boxed in cyan) comes to a complete stop at $t=1$s before making a right turn. Similarly, a southbound vehicle (boxed in orange) halts at $t=7$s before proceeding, even though no other vehicles are entering the intersection at the same time. This strict compliance with traffic rules is achieved by transferring explicit rules from the raw scenario into the SUMO network. Furthermore, the simulation of parking behavior results from the enhanced control logic implemented in SUMO.

\subsubsection{Scenario 3: Simulating Diverse Driving Behaviors}

\begin{figure}[ht]
    \centering
    \includegraphics[width=1\linewidth]{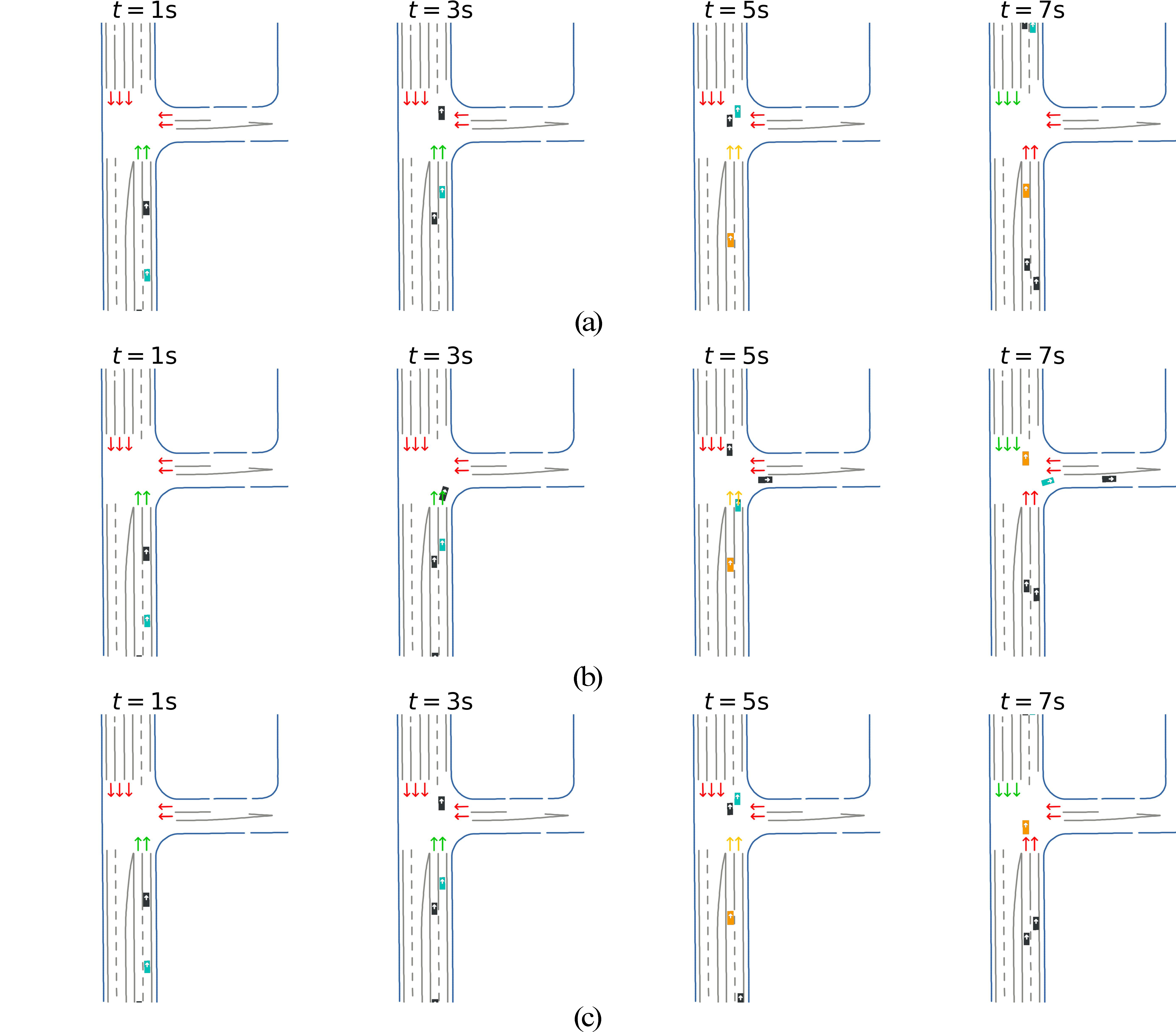}
    \caption{Demonstration of scenario 3. (a) Rollout 1, (b) Rollout 2, (c) Rollout 3.}
    \label{fig:scenario3}
\end{figure}

Our method also introduces randomness, enabling the simulation of diverse driving styles and routing decisions from the same initial context. This scenario (ID: 19a486cd29abd7a7) explores such diversity. Figure \ref{fig:scenario3} illustrates three simulation rollouts, labeled (a), (b), and (c). Two vehicles are highlighted as vehicles of interest.

In this three-way signalized intersection, the southbound signals transition from green to red at around $t=5\text{s}$, with a yellow light phase lasting approximately 2 seconds. This “dilemma zone” provides room for distinct decisions by the vehicle boxed in cyan. In rollouts (b) and (c), the vehicle maintains its speed and passes through the intersection during the yellow light. In contrast, in rollout (a), the vehicle decelerates to a complete stop, opting to wait for the next green light.

Additionally, the rightmost lane on the southbound approach allows both straight-through and right-turn movements, giving rise to different route choices for the vehicle boxed in orange. In rollouts (a) and (c), the vehicle continues straight on the main road, while in rollout (b), it slows down and turns right onto the branch road.

%% file: sections/results_wosac.tex
\subsection{Performance Evaluation for Short-Time Horizon Simulation Using WOSAC}
\label{sec:results-short-time-performance}

Waymo Open Sim Agents Challenge adopts three groups of metrics to collectively evaluate the realism of trajectory simulation:

\begin{itemize}
    \item \textbf{Kinematic metrics}, including linear speed, linear acceleration, angular speed, and angular acceleration, capture the fidelity of agent motions;

    \item \textbf{Interactive metrics}, including collision indication, distance to nearest object, and time to collision (TTC), capture the interaction of agents with each other;

    \item \textbf{Map-based metrics}, including offroad indication and distance to road edge, capture the agents' degree of adherence to the map.
\end{itemize}

These metrics are computed per scenario and averaged with weights to come up with a realism-meta metric ranging from 0 to 1, as a quantitative evaluation of the simulation model \cite{montali2024waymo-WOSAC-benchmark}.\footnote{For more information of WOSAC, please refer to \url{https://waymo.com/open/challenges/2024/sim-agents/}.}

\begin{table*}[t]
\centering
\caption{Comparison with state-of-the-art models on WOMD 2024 Sim Agents benchmark.}
\begin{tabular}{c|c|ccc|c}
\toprule
\multirow{2}{*}{\textbf{Method}} & \textbf{Realism} & \textbf{Kinematic} & \textbf{Interactive} & \textbf{Map-based} & \multirow{2}{*}{\textbf{minADE↓}} \\ 
& \textbf{Meta metric↑} & \textbf{metrics↑} & \textbf{metrics↑} & \textbf{metrics↑} & \\
\midrule
% SMART 96M       & \textbf{0.7564}               & 0.4768                      & 0.7986                        & \textbf{0.8618}             & 1.5500           \\
SMART 8M~\cite{wu2024smart}        & 0.7511                        & 0.4445                      & 0.8053& 0.8571                      & 1.5435           \\
BehaviorGPT~\cite{zhou2024behaviorgpt-behavior-model}     & 0.7473                        & 0.4333                      & 0.7997                        & 0.8593                      & 1.4147\\
GUMP~\cite{hu2024solving}            & 0.7431                        & 0.4780& 0.7887                        & 0.8359                      & 1.6041           \\
MVTE~\cite{wang2023multiverse}            & 0.7302                        & 0.4503                      & 0.7706                        & 0.8381                      & 1.6770           \\
VBD~\cite{huang2024versatile}             & 0.7200                        & 0.4169                      & 0.7819                        & 0.8137                      & 1.4743           \\
TrafficBots v1.5~\cite{zhang2024trafficbotsv15trafficsimulation}  & 0.6988                        & 0.4304                      & 0.7114                        & 0.8360                      & 1.8825           \\ 
\midrule
SUMO~\cite{krajzewicz2012sumo} & 0.6532 & 0.3294 & 0.7153 & 0.7585 & 5.8305 \\
\bottomrule
\end{tabular}
\label{tab:comparison-wosac}
\end{table*}

\begin{figure}
    \centering
    \includegraphics[width=1\linewidth]{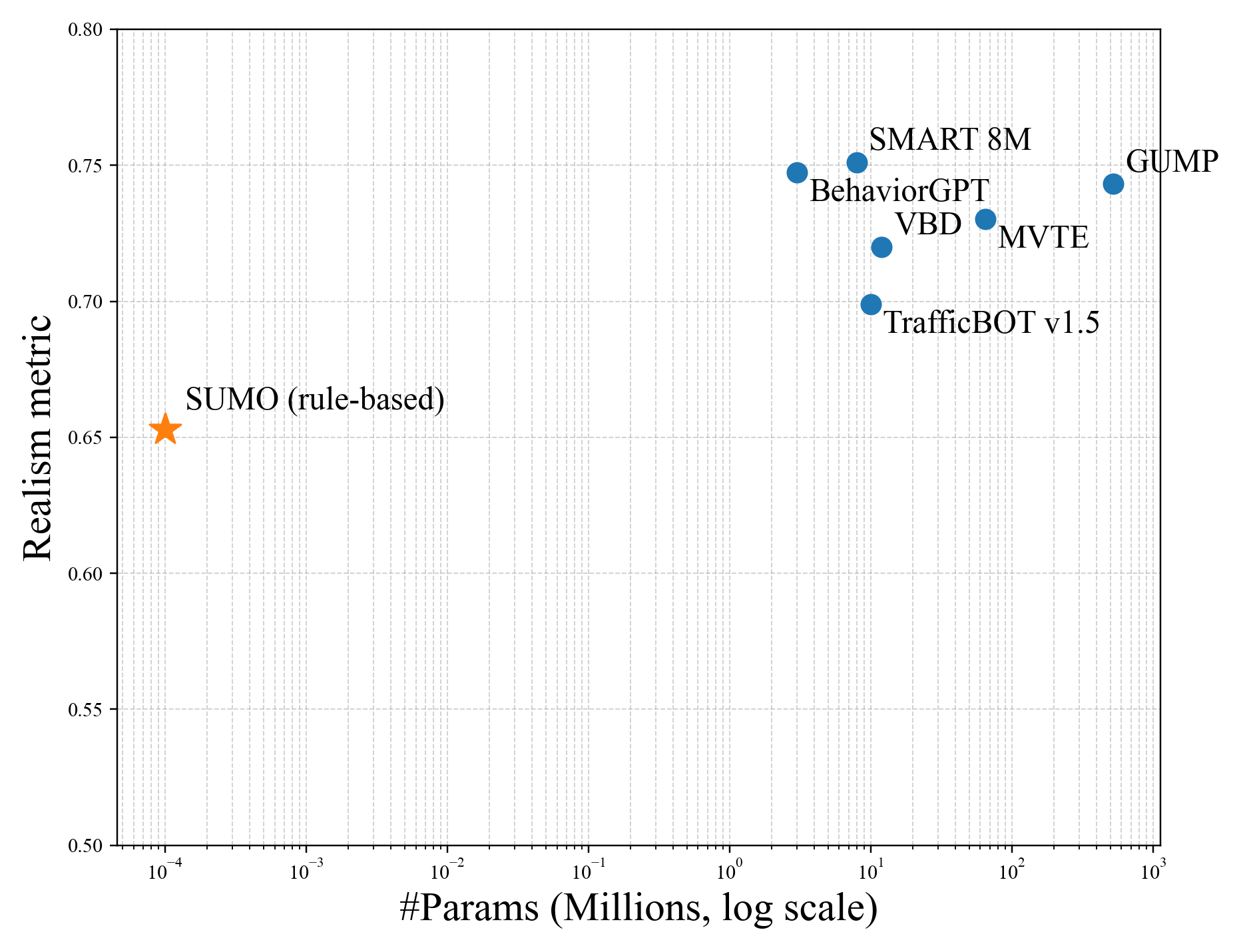}
    \caption{Relationship between model size and the WOSAC realism metric for SUMO and state-of-the-art data-driven simulators. Using only about 0.001\% of the model size of the best-performing method, SUMO achieves 86.9\% of its realism score.}
    \label{fig:params_vs_realism}
\end{figure}

Table \ref{tab:comparison-wosac} shows the metrics of test set on 2024 WOSAC. As shown, SUMO achieves a realism metric of 0.653. Though not comparable to the state-of-the-art learning-based simulation models, a profound advantage of SUMO as a model-based simulator over other existing learning-based, data-driven methods lies in its model scale, interpretability, computational burden, and efficiency. Model-based simulator requires much fewer parameters and less computational resources compared to other data-driven models. Specifically, only less than 100 tunable parameters are required in SUMO. Figure \ref{fig:params_vs_realism} reveals the scatter plot comparing the size and performance of SUMO to other existing models. It achieves 86.9\% of SOTA performance, with only 0.001\% of model parameters. As seen, SUMO is one of the most light-weighted simulators.

%% file: sections/results_long.tex
\subsection{Performance Evaluation for Long-Time Horizon Simulation}
\label{sec:results-long-time-performance}

While WOSAC is one of the most recognized and widely used simulation benchmarks in autonomous driving research, it primarily measures the performance of models in short-term simulations. In real-world driving environments, multiple agents continuously interact with each other for a long-time horizon. It is critical that a simulator is robust under long-time simulations. To systematically evaluate and compare long-horizon robustness, we first design and report quantitative long-horizon metrics, and then present two representative failure-case scenario analyses from data-driven simulators to explain the observed trends.

\subsubsection{Quantitative Evaluation with Long-Horizon Metrics}

\begin{table}[ht]
    \centering
    \caption{Comparison of long-time horizon metrics between SUMO, SMART, and TrafficBots v1.5.}
    \begin{tabular}{c|ccc}
    \toprule
        \textbf{Method} & \textbf{Collision Rate↓} & \textbf{Offroad Rate↓} \\
        \midrule
        SMART 8M~\cite{wu2024smart} & 0.0035 & 0.0206 \\
        TrafficBots v1.5~\cite{zhang2024trafficbotsv15trafficsimulation} & 0.2507 & 0.1526 \\
        SUMO~\cite{krajzewicz2012sumo} & 0.0047 & 0.0073 \\
        \bottomrule
    \end{tabular}
    \label{tab:longhorizonmetric}
\end{table}

To provide a quantitative comparison, we follow the same basic metric design philosophy from WOSAC, but extend the evaluation to 60 seconds. Concretely, we compute the collision rate and offroad rate by rolling out each simulator for the extended horizon. This extension stresses simulators under sustained multi-agent interaction, where compounding errors and distribution shift are more likely to emerge. To ensure a fair comparison, edge cases are carefully handled in the implementation, such as agents leaving the map area.

The quantitative comparison among SUMO, SMART, and TrafficBots is summarized in Table~\ref{tab:longhorizonmetric}. Overall, SUMO maintains low collision and offroad rates, indicating stable behavior generation and consistent map adherence. SMART also achieves a low collision rate, but exhibits a higher offroad rate than SUMO. In contrast, TrafficBots shows substantially higher collision and offroad rates, suggesting potential instability and compounding-error behaviors.

\subsubsection{Scenario 4 Simulated By SMART}
\begin{figure}[h!]
    \centering
    \includegraphics[width=1\linewidth]{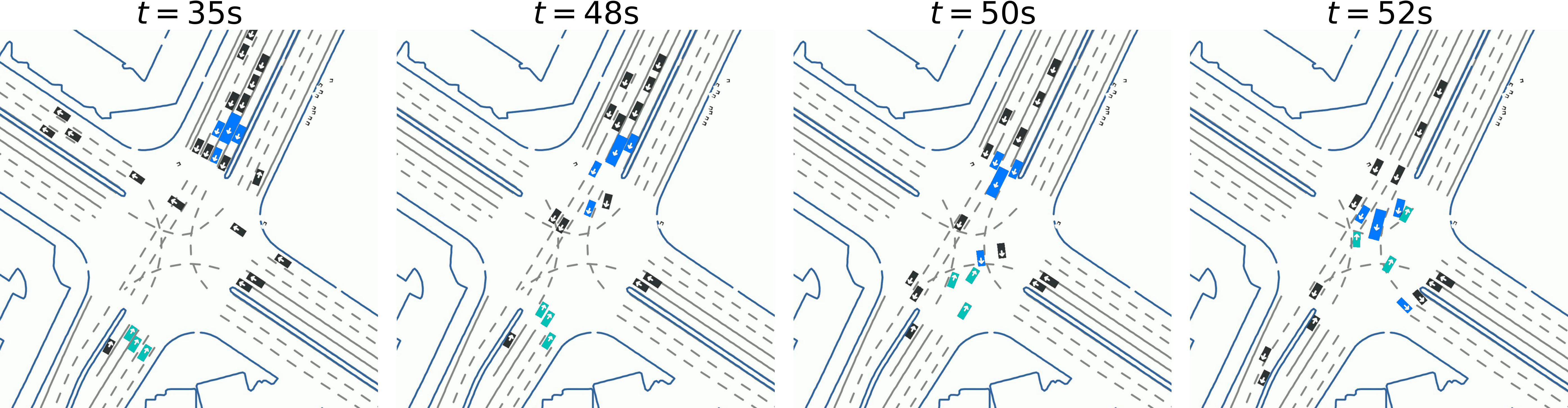}
    \caption{Scenario f4be7b9708a6d09 simulated by SMART.}
    \label{fig:scenario4}
\end{figure}

The simulation result for the first sample scenario (ID: f4be7b9708a6d09) is generated using SMART \cite{wu2024smart}, the state-of-the-art data-driven simulation model that achieved the highest realism metric score in WOSAC. Figure \ref{fig:scenario4} illustrates four time instances during a 6-second simulation generated by SMART's 8M parameter model. This scenario features a large signalized intersection. Vehicles highlighted in blue are approaching from the northbound, while vehicles highlighted in cyan are from the southbound.

Before $t=48$s, vehicles in both northbound and southbound directions are stationary, which is reasonable as eastbound vehicles are passing through the intersection. Beginning from $t=48$s, the eastbound vehicles clear the intersection, and both northbound and southbound vehicles, including left-turning and straight-going vehicles, begin to accelerate. However, the left-turning vehicles fail to yield the straight-going vehicles from the opposite direction. Between $t=50$s and $t=52$s, this lack of yielding results in multiple collisions within the intersection.

\subsubsection{Scenario 5 Simulated by TrafficBots}
\begin{figure}[h!]
    \centering
    \includegraphics[width=1\linewidth]{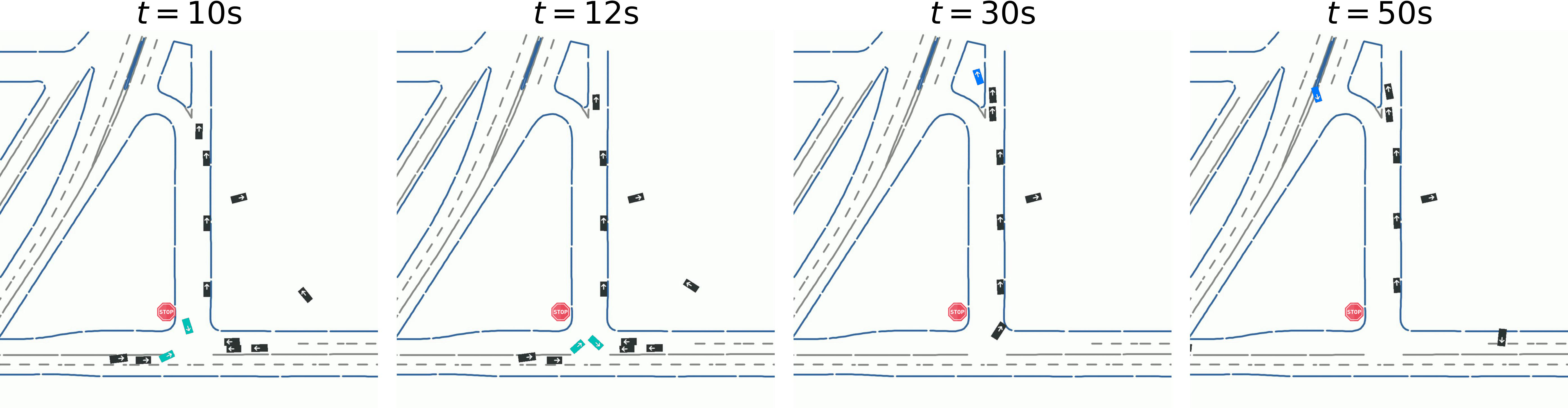}
    \caption{Scenario a60e78a32e03814f simulated by TrafficBotsV1.5.}
    \label{fig:scenario5}
\end{figure}

The simulation result for the second sample scenario (ID: a60e78a32e03814f), as shown in Figure \ref{fig:scenario5}, is generated by TrafficBotsV1.5 \cite{zhang2023trafficbots, zhang2024trafficbotsv15trafficsimulation}, another data-driven model. This scenario involves a stop-sign-controlled three-way intersection at the bottom of the map, where vehicles approaching from the northbound side road are required to stop and yield to vehicles on the main road. Beginning from $t=10$s, two vehicles highlighted in cyan approach this intersection. The one coming from the side road intends to turn left into the main road, while the one coming from the main road intends to turn left into the side road. However, the former does not stop and yield the latter, directly violating the stop-sign rule. They ended up turning within the intersection simultaneously.

In addition, the vehicle highlighted in blue exhibits unreasonable off-road behaviors from $t=30$s to $t=50$s. At $t=30$s, this vehicle veers off the road and enters an un-drivable area. By $t=50$s, its trajectory deviates even more and becomes entirely misaligned with the road geometry, failing to follow any path.

%% file: sections/discussions.tex
%%%%%%%%%%%%%%%%%%%%%%%%%%%%%%%%%%%%%%%%%%%%%%%%%%%%%%%%%%%%%%%%%%%%%%%%%%
\section{Discussion and Comparison between Model-Based and Learning-based Simulators}
\label{sec:discussion}

\input{sections/discussions_limitations}

\input{sections/discussions_advantages}

\input{sections/discussions_future}

%% file: sections/discussions_limitations.tex
\subsection{Limitations of Model-Based Simulators}
\label{sec:discussion-limitations}

\begin{figure}[h!]
    \centering
    \includegraphics[width=0.9\linewidth]{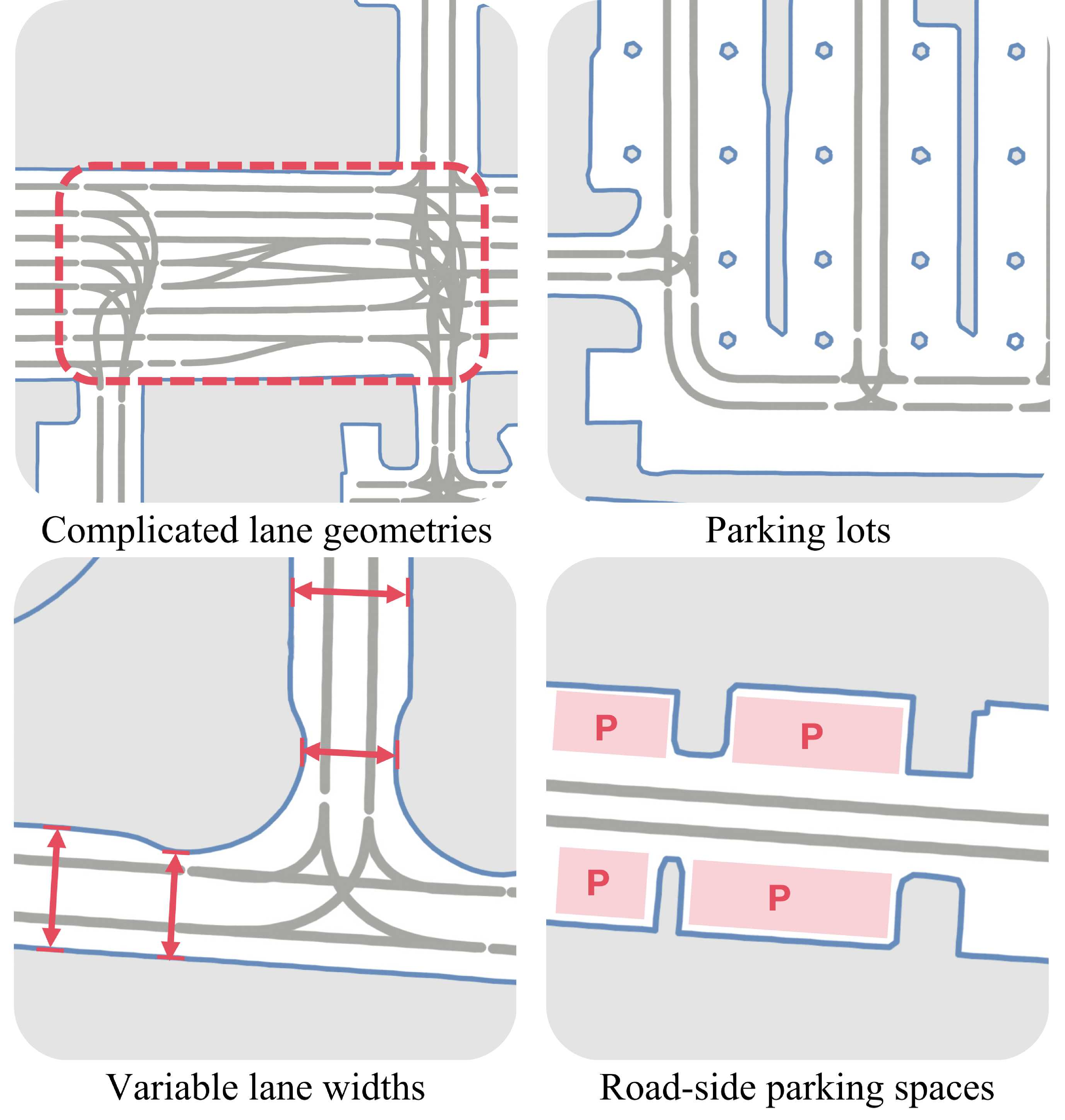}
    \caption{Examples of complex road structures that are unrepresentable by SUMO.}
    \label{fig:complex-road-structures}
\end{figure}

\subsubsection{Road Network-Level Limitations}
Model-based simulators like SUMO are built around structured, traffic-focused representations of road networks, which limits their ability to simulate more complex environments. Typical examples are depicted in Figure \ref{fig:complex-road-structures}.

\textbf{Heavy dependence on road network definition.} For model-based simulators, static contexts, including road networks, are fundamental to agent simulations. Agents are not allowed to be positioned in areas lacking network information. This dependency creates inflexibility, as it requires a complete and high-resolution road network for every area to be simulated. In contrast, data-driven models are more adaptable, allowing simulations to continue even in underinformed areas, although the realism of simulated agents in such regions may be reduced.

\textbf{Inability to represent complex road structures.} SUMO's road networks focus primarily on road segments relevant to traffic flow, but it fails to capture many intricate road structures seen in real-world environments. For instance, drivable areas such as driveways, parking lots, and roadside parking spaces cannot be represented. Additionally, SUMO's graph-based approach defines road segments through nodes and edges, as explained in Section \ref{sec:map-conversion}. This structure becomes unapplicable when dealing with complex geometries, such as successive merging or diverging lanes and dense intersections. This results in poor construction of road networks for these scenarios, degrading simulation quality. Another example lies in SUMO's inability to handle variable lane width and road boundaries. In SUMO, lane widths are fixed for a single lane and cannot dynamically adjust to match real-world road boundary conditions. As a result, when converting WOMD scenario to SUMO network, the road fringes in the SUMO network often fail to align with the original road boundaries from WOMD data, particularly in areas with irregular rugged roadbeds.

\subsubsection{Behavior-Level Limitations}

\begin{figure}[h!]
    \centering
    \includegraphics[width=0.9\linewidth]{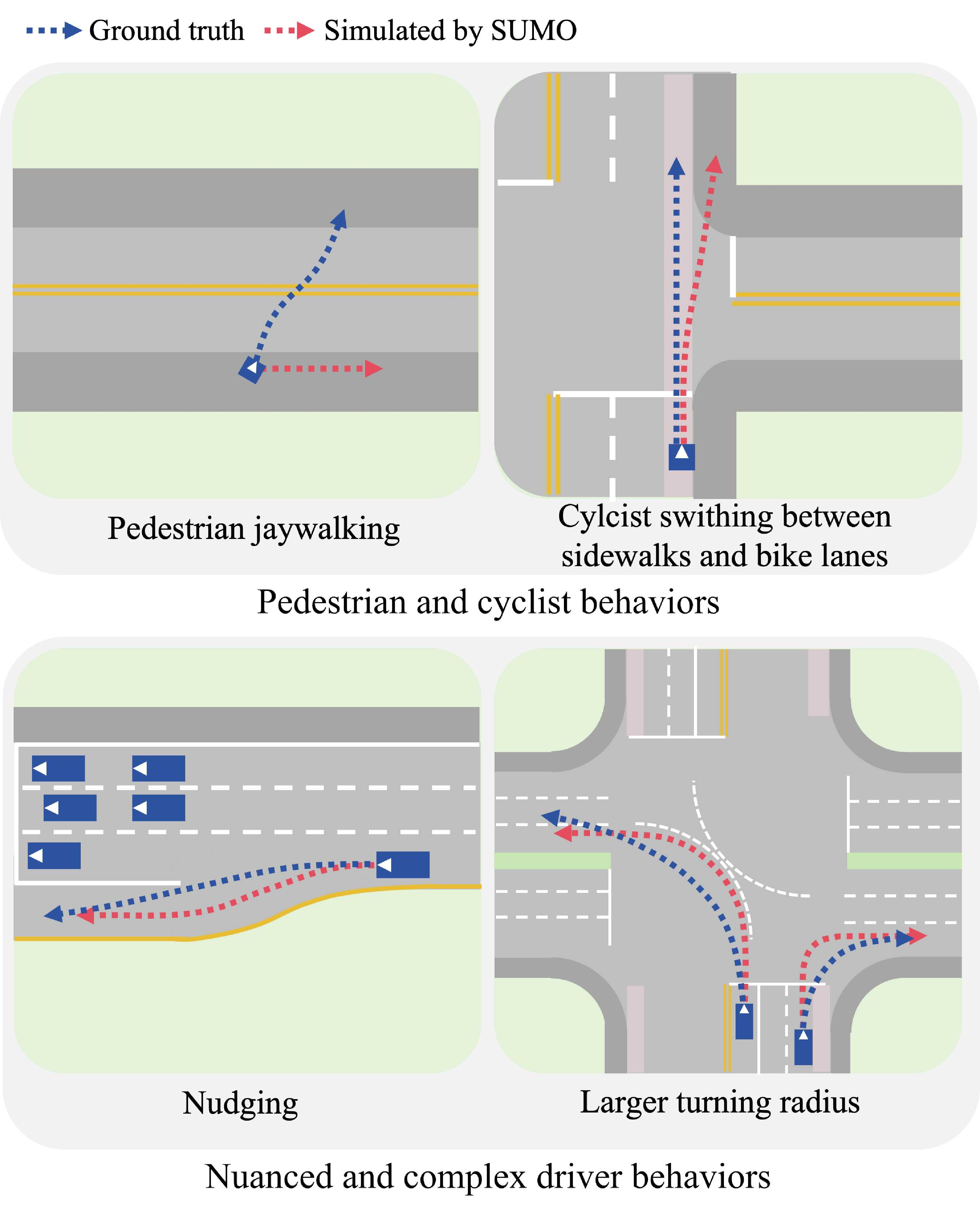}
    \caption{Examples of complex behaviors that are unproducible by SUMO.}
    \label{fig:behavior-level-limitations}
\end{figure}

\textbf{Pedestrian and Cyclist Behaviors.} While SUMO includes support for pedestrians and cyclists, their movements are confined to predefined sidewalks and crossings constructed within the network. This approach limits flexibility, as real-world walkable areas often extend beyond sidewalks to include parking lots, plazas, and open spaces. Pedestrian behavior is also more unpredictable compared to vehicle behavior; for example, pedestrians may jaywalk, stand idle, or behave erratically. Cyclists, similarly, may switch between roads and sidewalks or even ride against traffic. These irregular behaviors are difficult to simulate within SUMO's structured framework (Figure \ref{fig:behavior-level-limitations} top).

\textbf{Nuanced and Complex Driver Behaviors.} Human drivers exhibit subtle and complex behaviors that model-based simulators like SUMO cannot replicate. For example, nudging, where drivers inch forward in queues to signal intent, is challenging to simulate. Similarly, behaviors such as three-point turns, swerving, and micro-adjustments to maintain smoothness or avoid obstacles deviate from strict adherence to lane structures or rules. These micro-behaviors are beyond the capabilities of SUMO's rigid framework (Figure \ref{fig:behavior-level-limitations} bottom).

%% file: sections/discussions_advantages.tex
\subsection{Advantages of Model-Based Simulators}
\label{sec:discussion-advantages}

\textbf{Robustness.} Model-based simulators exhibit strong robustness in long-time horizon simulations and out-of-distribution scenarios, because they avoid the pitfalls of distribution shifts, issues commonly seen in data-driven models. In data driven models, small modelling errors may accumulate both in space and time, which might lead to out-of-distribution behaviors like frequent offroad, unrealistic collision, or even the collapse of the entire simulation.

\textbf{Interpretability and Controllability. }The decision-making processes within model-based simulators like SUMO are inherently interpretable, as they rely on explicitly defined rules and deterministic algorithms. For instance, lane-changing behavior in SUMO is governed by its lane-changing model, which evaluates factors such as route-following, speed optimization, and right-of-way adherence to determine vehicle actions at every simulation time step. In contrast, data-driven simulations learn and imitate patterns from training datasets, which makes their decision-making processes opaque. While the behavior of real-world human drivers is influenced by personal variability, it is fundamentally guided by clear and interpretable traffic rules, a characteristic well-represented in model-based simulators.

This interpretability leads directly to high controllability. In SUMO, vehicles can be precisely controlled via commands or parameter adjustments. For example, by modifying parameters such as lane-changing aggressiveness or vehicle acceleration desire, users can simulate diverse driving styles to fit specific needs, instead of doing random sampling from a distribution.

\textbf{Incorporation of Domain Knowledge. }Model-based simulators integrate well-established physical laws and expert knowledge, providing precise control over agent behavior. For example, SUMO's car-following and lane-changing models reflect principles from physical dynamics, so that the simulated vehicle trajectories are realistic in terms of physics. Another example is the ability of SUMO to establish and operate complete traffic light programs, which usually follow certain transportation industry standards. This enables model-based simulators to enforce adherence to traffic regulations.

\textbf{Generalizability. }Learning-based models often struggle with generalizability, under-performing on scenarios unseen in the training datasets. Instead, model-based simulators do not require extensive training. Once a static context conversion is completed, these simulators can adapt to new environments.

%% file: sections/discussions_future.tex
\subsection{Future directions}

Several research directions can build on this work and recent progress in world-model simulation:

Standardized long-horizon benchmarks. Extending community benchmarks toward longer horizons with clearly specified handling of invalid states would make robustness comparisons more reproducible.

Hybrid simulation that combines model-based constraints with learned generative agents. Recent controllable world-model and hybrid simulation frameworks aim to improve long-horizon reliability and controllability, suggesting a promising path to integrate rule/geometry consistency from microscopic simulators with the realism of learned rollouts.

Richer semantics and broader agent coverage in scenario conversion. Improving support for complex drivable areas (e.g., parking lots/driveways) and more realistic pedestrian ad cyclist behaviors would reduce the representational gap between real-world datasets and structured traffic simulators, and enable more comprehensive closed-loop evaluation.

%% file: sections/conclusions.tex
\section{Conclusions}
\label{sec:conclusion}

We introduced Waymo2SUMO, an automated pipeline that converts WOMD scenarios into SUMO by constructing road networks, estimating traffic signals, and initializing multi-agent dynamics with engineered control logic for real-world edge cases. This enables, to our knowledge, the first systematic large-scale benchmarking of a widely used model-based microscopic simulator on WOMD/WOSAC. Quantitative results show that SUMO achieves a 0.653 short-horizon realism meta metric on WOSAC while remaining extremely lightweight, and maintains strong long-horizon robustness in 60-second rollouts with low collision and offroad rates. Together, the findings clarify practical trade-offs between model-based and data-driven simulators and motivate future hybrid approaches that combine long-horizon stability and rule consistency with the high short-horizon realism of learned models.

%% file: ref.bib
@techreport{koopman2018framework,
  title        = {Toward a Framework for Highly Automated Vehicle Safety Validation},
  author       = {Philip Koopman and Michael Wagner},
  year         = {2018},
  institution  = {SAE International},
  number       = {SAE Technical Paper 2018-01-1071},
  doi          = {10.4271/2018-01-1071}
}

@misc{waymo_service_areas,
  author = {{Waymo}},
  title  = {Service areas},
  year   = {2025},
  note   = {Waymo Help Center}
}

@inproceedings{ettinger2021large-womd,
  title={Large scale interactive motion forecasting for autonomous driving: The waymo open motion dataset},
  author={Ettinger, Scott and Cheng, Shuyang and Caine, Benjamin and Liu, Chenxi and Zhao, Hang and Pradhan, Sabeek and Chai, Yuning and Sapp, Ben and Qi, Charles R and Zhou, Yin and others},
  booktitle={Proceedings of the IEEE/CVF International Conference on Computer Vision},
  pages={9710--9719},
  year={2021}
}

@article{montali2024waymo-WOSAC-benchmark,
  title={The {Waymo} open sim agents challenge},
  author={Montali, Nico and Lambert, John and Mougin, Paul and Kuefler, Alex and Rhinehart, Nicholas and Li, Michelle and Gulino, Cole and Emrich, Tristan and Yang, Zoey and Whiteson, Shimon and others},
  journal={Advances in Neural Information Processing Systems},
  volume={36},
  year={2024}
}

@inproceedings{caesar2020nuscenes-NuScenes,
  title={nuscenes: A multimodal dataset for autonomous driving},
  author={Caesar, Holger and Bankiti, Varun and Lang, Alex H and Vora, Sourabh and Liong, Venice Erin and Xu, Qiang and Krishnan, Anush and Pan, Yu and Baldan, Giancarlo and Beijbom, Oscar},
  booktitle={Proceedings of the IEEE/CVF conference on computer vision and pattern recognition},
  pages={11621--11631},
  year={2020}
}

@article{huang2024versatile,
  title={Versatile behavior diffusion for generalized traffic agent simulation},
  author={Huang, Zhiyu and Zhang, Zixu and Vaidya, Ameya and Chen, Yuxiao and Lv, Chen and Fisac, Jaime Fern{\'a}ndez},
  journal={arXiv preprint arXiv:2404.02524},
  year={2024}
}

@article{zhou2024behaviorgpt-behavior-model,
  title={BehaviorGPT: Smart Agent Simulation for Autonomous Driving with Next-Patch Prediction},
  author={Zhou, Zikang and Hu, Haibo and Chen, Xinhong and Wang, Jianping and Guan, Nan and Wu, Kui and Li, Yung-Hui and Huang, Yu-Kai and Xue, Chun Jason},
  journal={arXiv preprint arXiv:2405.17372},
  year={2024}
}

@article{zhong2021survey,
  title={A survey on scenario-based testing for automated driving systems in high-fidelity simulation},
  author={Zhong, Ziyuan and Tang, Yun and Zhou, Yuan and Neves, Vania de Oliveira and Liu, Yang and Ray, Baishakhi},
  journal={arXiv preprint arXiv:2112.00964},
  year={2021}
}

@misc{gulino2023waymax,
  title        = {Waymax: An Accelerated, Data-Driven Simulator for Large-Scale Autonomous Driving Research},
  author       = {Cole Gulino and Justin Fu and Wenjie Luo and George Tucker and Eli Bronstein and Yiren Lu and Jean Harb and Xinlei Pan and Yan Wang and Xiangyu Chen and others},
  year         = {2023},
  howpublished = {arXiv preprint arXiv:2310.08710}
}

@misc{caesar2021nuplan,
  title        = {nuPlan: A closed-loop ML-based planning benchmark for autonomous vehicles},
  author       = {Holger Caesar and Juraj Kabzan and Kok Seang Tan and Whye Kit Fong and Eric Wolff and Alex Lang and Luke Fletcher and Oscar Beijbom and Sammy Omari},
  year         = {2021},
  howpublished = {arXiv preprint arXiv:2106.11810}
}

@incollection{fellendorf2010vissim,
  title     = {Microscopic Traffic Flow Simulator VISSIM},
  author    = {Martin Fellendorf and Peter Vortisch},
  booktitle = {Fundamentals of Traffic Simulation},
  editor    = {Jaume Barcel{\'o}},
  publisher = {Springer},
  year      = {2010}
}

@incollection{casas2010aimsun,
  title     = {Traffic Simulation with Aimsun},
  author    = {Jordi Casas and Jaime L. Ferrer and David Garcia and Josep Perarnau and Alex Torday},
  booktitle = {Fundamentals of Traffic Simulation},
  editor    = {Jaume Barcel{\'o}},
  publisher = {Springer},
  year      = {2010}
}

@inproceedings{dosovitskiy2017carla,
  title     = {CARLA: An Open Urban Driving Simulator},
  author    = {Alexey Dosovitskiy and German Ros and Felipe Codevilla and Antonio Lopez and Vladlen Koltun},
  booktitle = {Proceedings of the Conference on Robot Learning (CoRL)},
  year      = {2017}
}

@inproceedings{rong2020lgsvl,
  title     = {LGSVL Simulator: A High Fidelity Simulator for Autonomous Driving},
  author    = {Guodong Rong and Boyu Shin and Yujiao Xu and Wenjing Lin and Ziyu Chen and Siyuan Zhang and Yaqin Hu and Jingwen Zhang and Yichen Wei and Shaojie Yu and others},
  booktitle = {IEEE Intelligent Transportation Systems Conference (ITSC)},
  year      = {2020}
}

@misc{shah2017airsim,
  title       = {AirSim: High-Fidelity Visual and Physical Simulation for Autonomous Vehicles},
  author      = {Shital Shah and Debadeepta Dey and Chris Lovett and Ashish Kapoor},
  year        = {2017},
  howpublished = {arXiv preprint arXiv:1705.05065}
}

@article{wu2024smart,
  title     = {SMART: Scalable Multi-agent Real-time Simulation via Next-token Prediction},
  author    = {Wu, Wei and Feng, Xiaoxin and Gao, Ziyan and Kan, Yuheng},
  journal   = {arXiv preprint arXiv:2405.15677},
  year      = {2024},
}

@inproceedings{zhang2023trafficbots,
  title = {TrafficBots: Towards World Models for Autonomous Driving Simulation and Motion Prediction},
  booktitle = {International Conference on Robotics and Automation (ICRA)},
  author = {Zhang, Zhejun and Liniger, Alexander and Dai, Dengxin and Yu, Fisher and Van Gool, Luc},
  year = {2023},
}

@misc{zhang2024trafficbotsv15trafficsimulation,
      title={TrafficBots V1.5: Traffic Simulation via Conditional VAEs and Transformers with Relative Pose Encoding}, 
      author={Zhejun Zhang and Christos Sakaridis and Luc Van Gool},
      year={2024},
      eprint={2406.10898},
      archivePrefix={arXiv},
      primaryClass={cs.RO},
      url={https://arxiv.org/abs/2406.10898}, 
}

@misc{suo2021trafficsim,
      title={TrafficSim: Learning to Simulate Realistic Multi-Agent Behaviors}, 
      author={Simon Suo and Sebastian Regalado and Sergio Casas and Raquel Urtasun},
      year={2021},
      eprint={2101.06557},
      archivePrefix={arXiv},
      primaryClass={cs.RO},
      url={https://arxiv.org/abs/2101.06557}, 
}

@misc{waymo_simulationcity,
  author = {{Waymo}},
  title  = {Simulation City: Introducing Waymo's most advanced simulation system yet for autonomous driving},
  year   = {2021},
  note   = {Waypoint (Waymo Blog)}
}

@misc{nvidia_av_simulation,
  author = {{NVIDIA}},
  title  = {Autonomous Vehicle Simulation},
  year   = {2025},
  note   = {NVIDIA Developer}
}

@misc{waabi_world,
  author = {{Waabi}},
  title  = {Welcome to Waabi World},
  year   = {2022},
  note   = {Waabi Insights}
}

@inproceedings{yuan2021agentformer,
  title     = {AgentFormer: Agent-Aware Transformers for Socio-Temporal Multi-Agent Forecasting},
  author    = {Ye Yuan and Xinshuo Weng and Yanglan Ou and Kris Kitani},
  booktitle = {ICCV},
  year      = {2021}
}

@inproceedings{jiang2023motiondiffuser,
  title     = {MotionDiffuser: Controllable Multi-Agent Motion Prediction Using Diffusion},
  author    = {Chiyu Max Jiang and others},
  booktitle = {CVPR},
  year      = {2023}
}

@article{jiang2024scenediffuser,
  title={Scenediffuser: Efficient and controllable driving simulation initialization and rollout},
  author={Jiang, Max and Bai, Yijing and Cornman, Andre and Davis, Christopher and Huang, Xiukun and Jeon, Hong and Kulshrestha, Sakshum and Lambert, John and Li, Shuangyu and Zhou, Xuanyu and others},
  journal={Advances in Neural Information Processing Systems},
  volume={37},
  pages={55729--55760},
  year={2024}
}

@inproceedings{igl2022symphony,
  title={Symphony: Learning realistic and diverse agents for autonomous driving simulation},
  author={Igl, Maximilian and Kim, Daewoo and Kuefler, Alex and Mougin, Paul and Shah, Punit and Shiarlis, Kyriacos and Anguelov, Dragomir and Palatucci, Mark and White, Brandyn and Whiteson, Shimon},
  booktitle={2022 International Conference on Robotics and Automation (ICRA)},
  pages={2445--2451},
  year={2022},
  organization={IEEE}
}

@article{xu2022bits,
  title={Bits: Bi-level imitation for traffic simulation},
  author={Xu, Danfei and Chen, Yuxiao and Ivanovic, Boris and Pavone, Marco},
  journal={arXiv preprint arXiv:2208.12403},
  year={2022}
}

@inproceedings{hu2024solving,
  title={Solving motion planning tasks with a scalable generative model},
  author={Hu, Yihan and Chai, Siqi and Yang, Zhening and Qian, Jingyu and Li, Kun and Shao, Wenxin and Zhang, Haichao and Xu, Wei and Liu, Qiang},
  booktitle={European Conference on Computer Vision},
  pages={386--404},
  year={2024},
  organization={Springer}
}

@article{wang2023multiverse,
  title={Multiverse transformer: 1st place solution for waymo open sim agents challenge 2023},
  author={Wang, Yu and Zhao, Tiebiao and Yi, Fan},
  journal={arXiv preprint arXiv:2306.11868},
  year={2023}
}

@article{krajzewicz2012sumo,
  title   = {Recent Development and Applications of SUMO -- Simulation of Urban MObility},
  author  = {Daniel Krajzewicz and Jakob Erdmann and Michael Behrisch and Laura Bieker},
  journal = {International Journal On Advances in Systems and Measurements},
  year    = {2012},
  volume  = {5},
  number  = {3 \& 4},
  pages   = {128--138}
}

@InProceedings{ermann2014sumo-intersection,
author="Erdmann, Jakob
and Krajzewicz, Daniel",
editor="Behrisch, Michael
and Krajzewicz, Daniel
and Weber, Melanie",
title="SUMO's Road Intersection Model",
booktitle="Simulation of Urban Mobility",
year="2014",
publisher="Springer Berlin Heidelberg",
address="Berlin, Heidelberg",
pages="3--17",
isbn="978-3-662-45079-6"
}

@InProceedings{erdmann2015-sumo-lanechange,
author="Erdmann, Jakob",
editor="Behrisch, Michael
and Weber, Melanie",
title="SUMO's Lane-Changing Model",
booktitle="Modeling Mobility with Open Data",
year="2015",
publisher="Springer International Publishing",
address="Cham",
pages="105--123",
isbn="978-3-319-15024-6"
}

@article{krauss1997-sumo-carfollowing,
  title = {Metastable states in a microscopic model of traffic flow},
  author = {Krauss, S. and Wagner, P. and Gawron, C.},
  journal = {Phys. Rev. E},
  volume = {55},
  issue = {5},
  pages = {5597--5602},
  numpages = {0},
  year = {1997},
  month = {May},
  publisher = {American Physical Society},
  doi = {10.1103/PhysRevE.55.5597},
  url = {https://link.aps.org/doi/10.1103/PhysRevE.55.5597}
}

@article{yan2026improving,
  title={Improving traffic signal data quality for the Waymo open motion dataset},
  author={Yan, Xintao and Liang, Erdao and Wang, Jiawei and Zhu, Haojie and Liu, Henry X},
  journal={Transportation Research Part C: Emerging Technologies},
  volume={183},
  pages={105476},
  year={2026},
  publisher={Elsevier}
}

@book{urbanik2015signal,
  title={Signal timing manual},
  author={Urbanik, Thomas and Tanaka, Alison and Lozner, Bailey and Lindstrom, Eric and Lee, Kevin and Quayle, Shaun and Beaird, Scott and Tsoi, Shing and Ryus, Paul and Gettman, Doug and others},
  volume={1},
  year={2015},
  publisher={Transportation Research Board Washington, DC}
}

@inproceedings{ross2011dagger,
  title     = {A Reduction of Imitation Learning and Structured Prediction to No-Regret Online Learning},
  author    = {Ross, St{\'e}phane and Gordon, Geoffrey J. and Bagnell, J. Andrew},
  booktitle = {Proceedings of the Fourteenth International Conference on Artificial Intelligence and Statistics (AISTATS)},
  pages     = {627--635},
  year      = {2011}
}

@inproceedings{poggenhans2018lanelet2,
  title     = {Lanelet2: A High-Definition Map Framework for the Future of Automated Driving},
  author    = {Poggenhans, Fabian and Pauls, Jan-Hendrik and Orf, Stefan and Naumann, Maximilian and Kuhnt, Florian and Mayr, Matthias},
  booktitle = {2018 21st International Conference on Intelligent Transportation Systems (ITSC)},
  year      = {2018},
  doi       = {10.1109/ITSC.2018.8569929}
}

@inproceedings{maierhofer2021commonroad,
  title     = {CommonRoad Scenario Designer: An Open-Source Toolbox for Map Conversion and Scenario Creation for Autonomous Vehicles},
  author    = {Maierhofer, Sebastian and Klischat, Moritz and Althoff, Matthias},
  booktitle = {2021 IEEE International Conference on Intelligent Transportation Systems (ITSC)},
  year      = {2021},
  doi       = {10.1109/ITSC48978.2021.9564885}
}

@inproceedings{lin2023openscenario2commonroad,
  title     = {Automatic Traffic Scenario Conversion from OpenSCENARIO to CommonRoad},
  author    = {Lin, Yuanfei and Ratzel, Michael and Althoff, Matthias},
  booktitle = {2023 IEEE 26th International Conference on Intelligent Transportation Systems (ITSC)},
  year      = {2023},
  doi       = {10.1109/ITSC57777.2023.10422422}
}

@inproceedings{vinitsky2022nocturne,
  title     = {Nocturne: A Scalable Driving Benchmark for Bringing Multi-Agent Learning One Step Closer to the Real World},
  author    = {Vinitsky, Eugene and Lichtl{\'e}, Nathan and Yang, Xiaomeng and Amos, Brandon and Foerster, Jakob},
  booktitle = {Advances in Neural Information Processing Systems (NeurIPS) Datasets and Benchmarks Track},
  year      = {2022}
}

@misc{li2021metadrive,
  title        = {MetaDrive: Composing Diverse Driving Scenarios for Generalizable Reinforcement Learning},
  author       = {Li, Quanyi and Peng, Zhenghao and Feng, Lan and Zhang, Qihang and Xue, Zhenghai and Zhou, Bolei},
  year         = {2021},
  howpublished = {arXiv preprint arXiv:2109.12674}
}

@inproceedings{majumdar2021paracosm,
  title     = {Paracosm: A Test Framework for Autonomous Driving Simulations},
  author    = {Majumdar, Rupak and Mathur, Arjun S. and others},
  booktitle = {Fundamental Approaches to Software Engineering (FASE 2021)},
  year      = {2021},
  doi       = {10.1007/978-3-030-71500-7_9}
}

@article{li2024choose,
  title={Choose your simulator wisely: A review on open-source simulators for autonomous driving},
  author={Li, Yueyuan and Yuan, Wei and Zhang, Songan and Yan, Weihao and Shen, Qiyuan and Wang, Chunxiang and Yang, Ming},
  journal={IEEE Transactions on Intelligent Vehicles},
  volume={9},
  number={5},
  pages={4861--4876},
  year={2024},
  publisher={IEEE}
}
